\documentclass[letterpaper, 10pt, conference]{ieeeconf}
\IEEEoverridecommandlockouts 
\makeatletter
\let\NAT@parse\undefined 
\makeatother
\usepackage[numbers,sort&compress]{natbib}
\usepackage{amsmath,amssymb,amsfonts}

\usepackage{amsthm}
\usepackage{algorithm}
\usepackage[noend]{algpseudocode}
\usepackage{float}
\usepackage{xcolor}
\usepackage{xspace}
\usepackage{graphicx}
\usepackage{booktabs}
\usepackage{array}
\usepackage{capt-of}
\usepackage{etoolbox}
\usepackage{url}
\usepackage{xurl}
\newlength{\tblwidth}
\newcolumntype{L}{l}                         
\newcolumntype{C}{c}
\newcolumntype{R}{r}

\providecommand{\credit}[1]{}
\def\tsc#1{\csdef{#1}{\textsc{\lowercase{#1}}\xspace}}
\tsc{WGM}
\tsc{QE}

\theoremstyle{plain}
\newtheorem{theorem}{Theorem}
\newtheorem{lemma}{Lemma}

\theoremstyle{definition}
\newtheorem{definition}{Definition}

\theoremstyle{remark}
\newtheorem{remark}{Remark}

\newcommand{\hidecomments}[1]{}

\makeatletter
\newcommand{\fs@insetruled}{%
  \fs@ruled
  \def\@fs@pre{\kern4pt\hrule height.8pt depth0pt\kern2pt}%
}
\makeatother
\floatstyle{insetruled}
\restylefloat{algorithm}

\newcommand{\ours}{{LSRP*}\xspace}
\newcommand{\flip}{\textsc{Flip}\xspace}

\newcommand{\dom}{\ensuremath{\succ}\xspace}
\newcommand{\nextevent}{\ensuremath{\tau_{\text{next}}(a)}\xspace}

\newcommand{\dur}[3]{D(#1,#2,#3)}
 
\begin{document}

\title{\LARGE \bf
LSRP*: Scalable and Anytime Planning for Multi-Agent Path Finding with Asynchronous Actions
}

\author{Dachuan Liu$^{1}$, Shuai Zhou$^{2}$, Shizhe Zhao$^{1}$,
and Zhongqiang Ren$^{1,\dagger}$
\thanks{$^{1}$Dachuan Liu, Shizhe Zhao, and Zhongqiang Ren are with Shanghai Jiao Tong
University, Shanghai, China {\tt\small dachuan@umich.edu,
\{shizhe.zhao, zhongqiang.ren\}@sjtu.edu.cn}}
\thanks{$^{2}$Shuai Zhou is with the South China University of Technology, Guangzhou,
China {\tt\small davidzhou718@gmail.com}}
\thanks{$^{\dagger}$Corresponding author.}
}

\maketitle
\thispagestyle{empty}
\pagestyle{empty}

\begin{abstract}
Multi-Agent Path Finding (MAPF) seeks collision-free paths for multiple agents from their respective starting locations to their respective goal locations while minimizing path costs.
Although many MAPF algorithms were developed, most of them rely on a common assumption on synchronized actions, where the actions of all agents start at the same time and always take a time unit.
This assumption may limit the use of MAPF planners in practice.
To get rid of this assumption, recently, an algorithm called Loosely Synchronized Rule-based Planning (LSRP) was proposed, which can find sub-optimal solutions for many agents. However, LSRP often finds poor quality solutions due to its unbounded sub-optimality. This paper develops a new anytime planner called LSRP* that can keep improving solution quality after the initial solution is obtained until the runtime budget depletes. We analyze the properties of LSRP* and test it against several baselines with up to 1000 agents on various maps. LSRP* can handle up to 25\% more agents than LSRP and can reduce up to 40\% of the solution cost found by LSRP.
\end{abstract}

\section{Introduction}\label{lsrp-star:intro}\label{lsrp-star-main:intro}
Multi-Agent Path Finding (MAPF) seeks collision-free paths for multiple agents from their start locations to goal locations while minimizing the path costs.
MAPF arises in applications such as logistics and automated warehouses.
The shared environment of the agents is often represented by a graph, where vertices represent the locations that the agent can reach, and edges represent the transition between two locations.
MAPF is NP-hard to solve to optimality \cite{yu2013structure}, and many MAPF planners were developed, ranging from unbounded sub-optimal planners~\cite{okumura2022priority,de2013push}, bounded sub-optimal planners~\cite{barer2014suboptimal,li2021eecbs} to optimal planners~\cite{sharon2015conflict,wagner2015subdimensional}, trading off completeness and solution optimality for scalability.

Most existing MAPF planners rely on a common assumption that all agents move in a synchronous manner: it takes the same amount of time for any agent to go through any edge, and all agents start and end each of their actions at the same time step.
However, this assumption limits the use of MAPF planners in practice, especially when agents have different action durations, such as different speeds or different edge lengths.
Consider mobile robots transporting materials in a factory, where a loaded agent has to traverse an edge slowly while an unloaded agent can move fast to traverse the same edge.

\stepcounter{footnote}
\begin{figure}[tb]
  \centering
  \includegraphics[width=\linewidth]{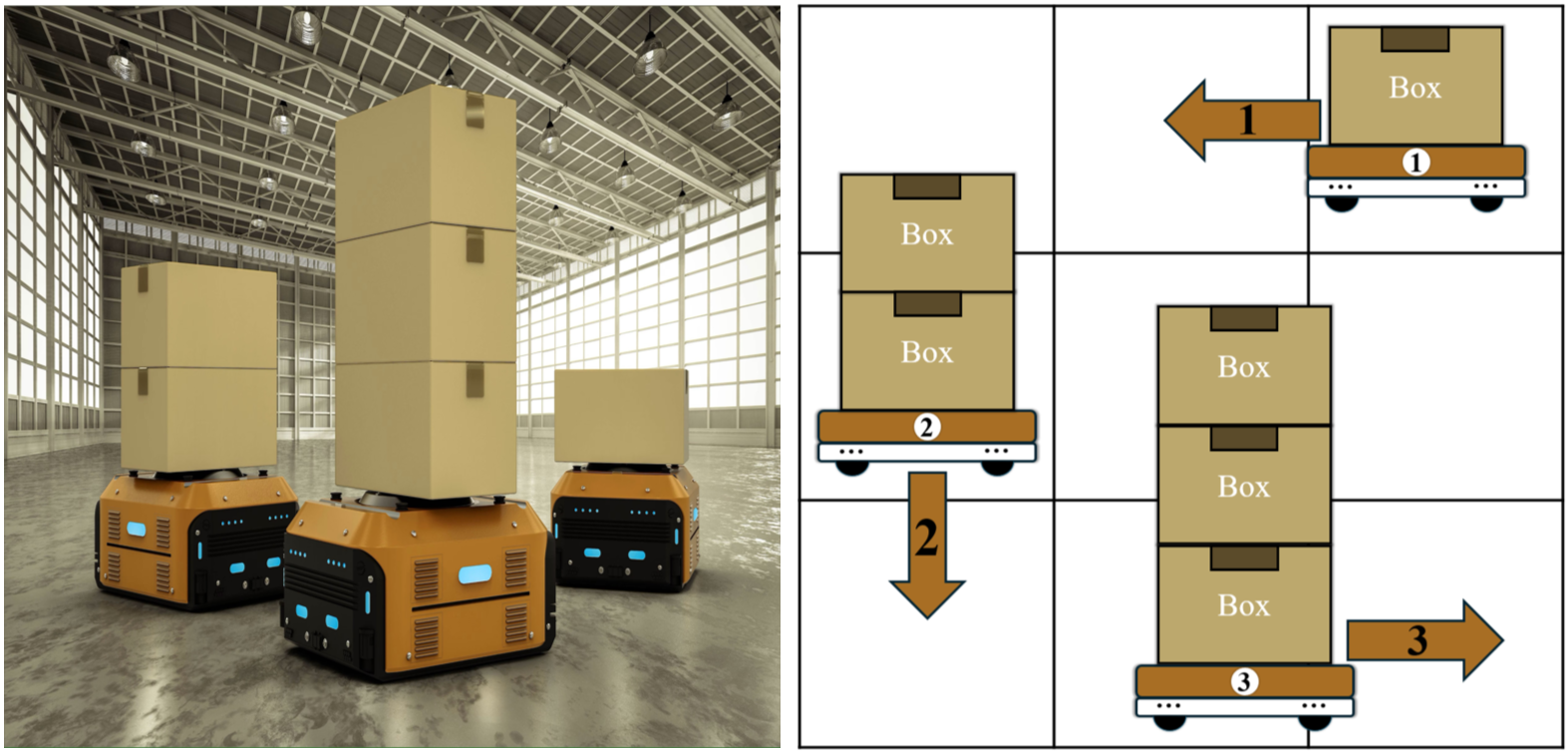}
  \caption{Motivation for MAPF-AA. Robots may have different speeds due to payload and
  safety reasons, which leads to the asynchronous actions as considered in this
  work.\protect\footnotemark[\value{footnote}]}
  \label{fig:mapfaa}
\end{figure}
\footnotetext[\value{footnote}]{\raggedright Figure source:
\url{https://www.vicorpower.com/resource-library/case-studies/warehouse-robot}}

To bypass this assumption on synchronous actions, a few variants of MAPF were studied, such as Continuous-Time MAPF~\cite{ANDREYCHUK2022103662}, MAPF with Asynchronous Actions (MAPF-AA)~\cite{ren2021loosely}, MAPF$_R$~\cite{walker2018extended}, and only a few algorithms were developed to address these problems.
Most of these algorithms seek an optimal or bounded sub-optimal solution at the cost of limited scalability as the number of agents grows. 
Recently, an unbounded sub-optimal
planner Loosely Synchronized Rule-based Planning (LSRP)~\cite{zhou2025loosely} for MAPF-AA was developed. Although LSRP can handle a large number of agents by using an iterative push and swap strategy as in PIBT~\cite{okumura2022priority}, LSRP often finds poor quality solutions due
to its unbounded sub-optimality. This paper thus develops LSRP*, an anytime variant of LSRP that can keep improving solution quality till a given runtime budget depletes.

In the literature on MAPF, LaCAM~\cite{okumura2023lacam} introduces a two-level search structure that generates the successors of a joint configuration one at a time, making the search over the joint configuration space complete and fast; LaCAM terminates once it finds a single collision-free joint path.
Its anytime extension LaCAM*~\cite{okumura2023improving} keeps searching after the first solution and iteratively refines the joint path toward the optimum as the runtime budget allows.
Adapting this two-level, anytime design from MAPF to LSRP for MAPF-AA, however, is challenging. First, the joint action space of all agents in MAPF-AA is complicated due to the different action durations, and one has to include the start and end times of those actions during planning since
the same joint action starting from different times may lead
to agent-agent collision. Second, when the search finds two
joint paths that reach the same joint configuration, simply
keeping the joint path with the smaller cost may not
eventually lead to a better solution, since the agents' arrival
times also matter and affect the potential agent-agent collision in the future path.

To address these challenges, LSRP* first searches a time-augmented state space and incorporates the agents' arrival times into the exploration process of the joint action space when refining the solution. Second, LSRP* introduces a new
pruning strategy to compare two joint paths that reach the same joint edge by considering both the cost and the arrival times of the agents. Finally, to further improve scalability and solution quality, LSRP* introduces a Flip mechanism based on simulated push delays of the agents during planning, to bypass traffic congestion in MAPF-AA.

With these techniques, we show that LSRP* is complete and guaranteed to eventually find an optimal solution given sufficient runtime.
In our tests, we compare LSRP* against several baselines, including LSRP \cite{zhou2025loosely}, CBS-AA \cite{wu2026conflict}, and prioritized planning with SIPP \cite{phillips2011sipp}, using various maps from a popular MAPF benchmark \cite{stern2019multi}. The results show that LSRP* can solve instances with up to 25 times more agents than these baselines and can refine the solution cost of the initial LSRP solution by up to about 35\% within a runtime budget, reaching within a few percent of the optimum on instances where an optimal solution can be computed.

\section{Related Work}\label{sec:related}
Our work is most related to LaCAM among search-based MAPF planners and to rule-based planning for MAPF-AA.

\textbf{LaCAM and its anytime extension.}
LaCAM~\cite{okumura2023lacam} is a complete, search-based planner for standard MAPF.
Its key idea is a two-level, lazy expansion: each high-level node holds a configuration and gradually grows a low-level constraint tree, generating successor configurations one at a time instead of all at once.
LaCAM*~\cite{okumura2023improving} then turns this search into an anytime, eventually optimal planner: it keeps expanding after the first solution so that the incumbent converges to the optimum as the runtime budget allows.
However, both of them assume synchronous actions, where all agents share a common time step. LSRP* is inspired by this design, but adapts it to MAPF-AA with a time-augmented state space and a dominance rule over both cost and arrival times. 
Besides LaCAM*, large neighborhood search
(MAPF-LNS~\cite{li2021anytime}, MAPF-LNS2~\cite{li2022mapf}) iteratively
replans subsets of agents to improve an initial solution. 
All these methods assume synchronous actions.

\textbf{Planning for MAPF-AA.}
Several variants of MAPF drop the synchronous-action assumption. MAPF$_R$ allows
non-unit edge costs and is solved optimally by E-ICTS~\cite{walker2018extended}.
Continuous-time CBS (CCBS)~\cite{ANDREYCHUK2022103662} extends CBS to continuous time
and uses SIPP~\cite{phillips2011sipp} as its low-level planner. For MAPF-AA,
Loosely Synchronized Search (LSS)~\cite{ren2021loosely} searches a time-augmented joint
state space and prunes joint states by dominance on the agents' arrival times, and
CBS-AA~\cite{wu2026conflict} and its bounded sub-optimal variant~\cite{2026_SOCS_BCBS_AA_XuemianWu} adapt CBS to asynchronous actions. These planners seek optimal or bounded sub-optimal solutions and were tested with up to tens of agents. 
For scalability, LSRP~\cite{zhou2025loosely} instead extends the rule-based pushing of PIBT~\cite{okumura2022priority} to asynchronous actions and scales to hundreds of agents, but it is unbounded sub-optimal and cannot use extra runtime to improve its solution.
LSRP* builds on LSRP and makes it anytime while retaining (and sometimes further improving) its scalability.

\section{Problem Definition}\label{lsrp-star:probDefine}\label{lsrp-star-main:problem-def}
Let $I =\{1,2,\dots,N\}$ denote a set of $N$ agents.
All agents move in a workspace modeled by a finite graph $G =(V,E)$,
where vertices $V$ represents all possible locations of the agents and 
edges $E\subseteq V\times V$ denote the pairs of adjacent vertices between which agents can move. For each agent $i$, let $\dur{i}{u}{v}\in\mathbb{R^+}$ be the required duration to move along edge $(u,v)\in E$. For different agents $i$ and $j$, their durations for the same edge may not be the same, i.e., $\dur{i}{u}{v} \neq \dur{j}{u}{v}$ is possible.

An \emph{action} is a tuple $a^i=(p,v,t_p,t_v)$. Throughout this paper, we use the \emph{parenthesis operator} to access a component of a tuple: given a tuple $y$ and one of its components $x$, we write $x(y)$ for that component of $y$. For example, $t_v(a^i)$ denotes the arrival time of action $a^i$, which we also write as $t^i_v$ when the action is clear from the context. Agent $i$ departs from vertex $p$ at time $t_p$ and arrives at vertex $v$ at time $t_v$ along the edge $(p,v)\in E$. It is a \emph{wait} if $p=v$, whose length $t_v-t_p$ is any positive real number, and a \emph{move} if $p\neq v$, whose length is $t_v-t_p=\dur{i}{p}{v}$. In both cases agent $i$ \emph{occupies}: (1) $p$ at $t_p$, (2) $v$ at $t_v$, (3) both $p,v$ within the open interval $(t_p,t_v)$. Two agents \emph{conflict} if they occupy the same vertex at the same time. We refer to this definition of conflict as the \emph{duration conflict} hereafter.

Let $v^i_s,v^i_g$ denote the start and goal vertex of agent $i$ respectively. All agents share a global clock and start their action at $v^i_s$ at time $t=0$.

A \emph{path} of agent $i$ is a sequence of consecutive actions
$\pi^i=\langle a^i_0,\dots,a^i_K\rangle$, i.e.,\ $v(a^i_k)=p(a^i_{k+1})$ and
$t_v(a^i_k)=t_p(a^i_{k+1})$, that starts at $v^i_s$ at time $0$ and ends at
$v^i_g$. For a prefix of $\pi^i$ ending at action $a^i_k$, its cost is
\begin{equation}
g(a^i_k) =
\begin{cases}
g(a^i_{k-1}), & k>0 \text{ and } p(a^i_k)=v(a^i_k)=v^i_g,\\
t_v(a^i_k), & \text{otherwise},
\end{cases}
\label{eq:cost}
\end{equation}
so the cost stays frozen while $i$ waits at its goal and grows again
once $i$ leaves it. The cost of the path is $g(\pi^i)=g(a^i_K)$.
A \emph{joint path} is
$\pi=(\pi^1,\dots,\pi^N)$; it is a \emph{solution} if no two agents conflict.
We denote $g(\pi)=\sum_{i=1}^N g(\pi^i)$ the \emph{sum of costs} (SOC) of a
solution.

Given $G$, the agents $I$ with their start and goal vertices,
and the durations $D$, the \emph{multi-agent path finding with asynchronous actions} (MAPF-AA) problem seeks a conflict-free joint path with minimum SOC. 
In contrast to classical MAPF, agents are not synchronized: the duration of a move depends on the agent and the edge, so there are different instants at which agents start new actions.

\section{Method}\label{lsrp-star:method}\label{lsrp-star-main:method}
\newcommand{\events}{\ensuremath{\Omega}\xspace}
\newcommand{\anext}{\ensuremath{\mathcal{A}_{\text{next}}}\xspace}
\newcommand{\adummy}{\ensuremath{a_{0}}\xspace}
\newcommand{\jv}{\ensuremath{\mathbf{v}}\xspace}

\subsection{Notations and Concepts}\label{method:notation}
\subsubsection{Events and Actions}
For a path $\pi^i$ of agent $i$, let $\events(\pi^i)$ denote the set of time
points at which the agent starts or finishes an action:
$$
\events(\pi^i) = \{t_p(a^i_k), t_v(a^i_k) \mid 0\leq k \leq K\},
$$
which denotes the set of \emph{events} of $\pi^i$. For a joint path $\pi$, let
$$
\events(\pi)=\bigcup_{i\in I} \events(\pi^i)
$$ denote the set of events of $\pi$.
Let $a=(a^i)_{i\in I}$ denote a joint action in $\pi$ at an event $t\in\events(\pi)$,
where the action of each agent is either in progress or completed at $t$.
Specifically, let $\tau(a)=\min\{t_v(a^i)\mid i\in I\}$ denote the earliest completion time among the actions of all agents, and let $I_{curr}(a)\subseteq I$ denote the set
of agents whose actions are completed at $\tau(a)$, other agent $j$ satisfies
$$
t_p(a^j)< \tau(a) <t_v(a^j) ,\quad \text{where } j\in I\setminus I_{curr}(a) .
$$
A joint path $\pi$ can be represented as a sequence of joint actions
$\pi=\langle a_0,\cdots,a_K\rangle$ with $\tau(a_0)<\tau(a_1)<\cdots<\tau(a_K)$.
We call $\tau(a_k)$ the \emph{current event} of $a_k$, and the earliest event after it
the \emph{next event}:
$$
\tau_{next}(a_k)=\min\{\,t\in\events(\pi)\mid t>\tau(a_k)\,\}=\tau(a_{k+1}),
$$
with $0\le k<K$.
During the search, the next event of $a$ is determined once its successor is generated as elaborated later in Sec.~\ref{method:preliminary:action tree}.
Let $\mathbf{v}=(v^1,\dots,v^N)$ denote the \emph{joint vertex} of $a$.
Let $e(a)=((p^1,v^1),\dots,(p^N,v^N))$ denote the \emph{joint edge} of $a$, and let $g(a)=\sum_{i\in I}g(a^i)$ denote its SOC.

In this work, we find the joint path $\pi$ by incrementally searching joint actions.
Specifically, for a given joint action $a$,
we explore successor joint actions $\anext(a)$, where each $a_{next}\in \anext(a)$
satisfies:
(i) every agent $i\in I_{curr}(a)$ starts a new action at $\tau(a)$;
(ii) every agent $j\in I\setminus I_{curr}(a)$ keeps its current unfinished action, i.e., $a^j_{next}=a^j$.
The search starts with a dummy joint action $\adummy$, in which each agent $i$
is at its start vertex $s^i$ at time $0$ with no action assigned yet, i.e.,
$a_0^i=(v_s^i,v_s^i,0,0)$, $\tau(\adummy)=0$ and $I_{curr}(\adummy)=I$.
We call agents in $I_{curr}$ \emph{idle agents} and others \emph{busy agents}. 

\begin{figure*}[tb]
  \centering
  \includegraphics[width=0.9\linewidth]{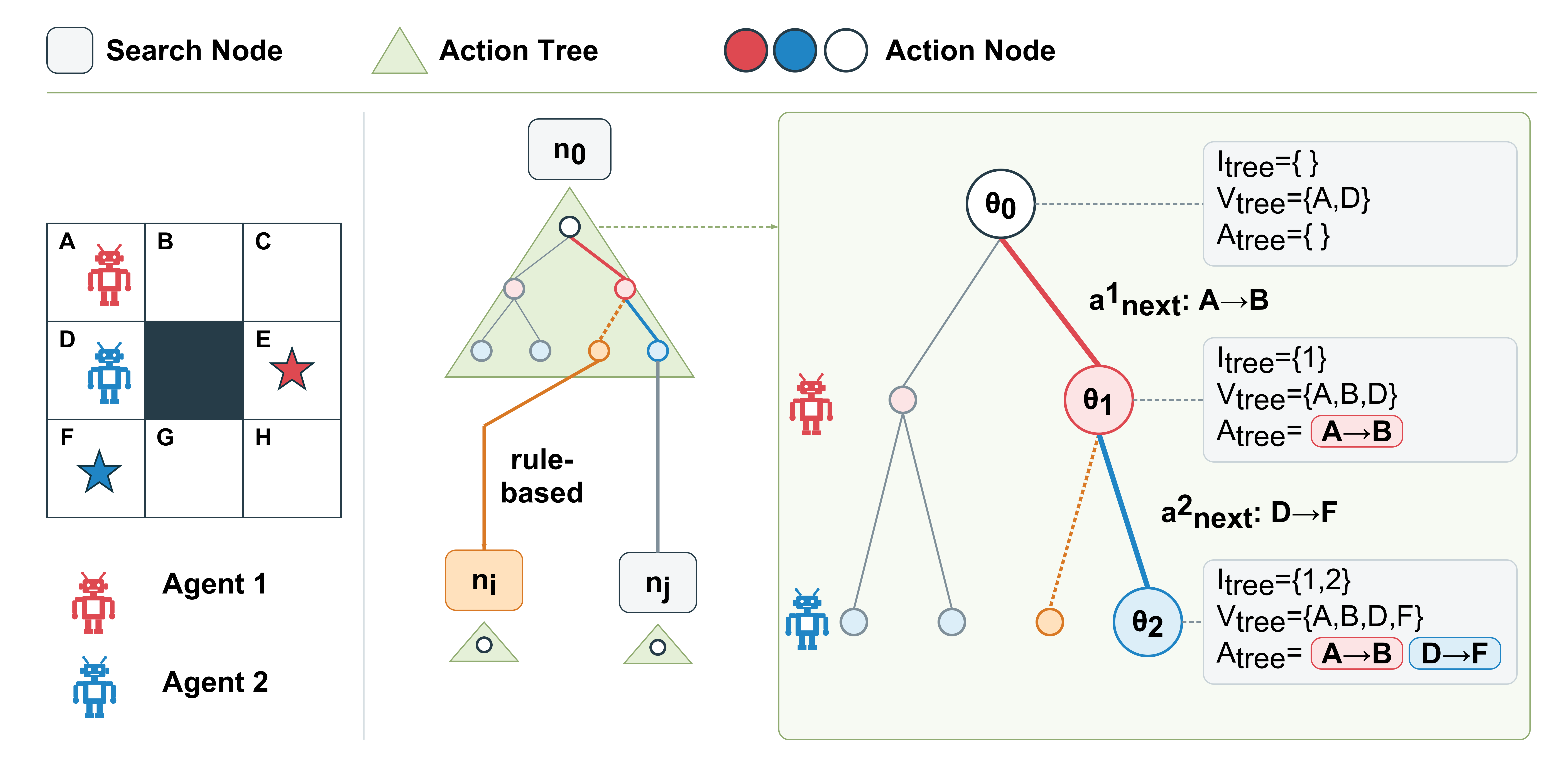}
  \caption{A toy example: each search node owns an action tree. The orange search
  node is generated from a partial assignment completed by the rule-based
  procedure; the gray one from a leaf that assigns every idle agent.}
  \label{fig:action_tree}
\end{figure*}

\subsubsection{Action tree}\label{method:preliminary:action tree}
For a joint action $a$, we enumerate its successors $\anext(a)$ via an
\emph{action tree} $AT(a)$. The action tree $AT(a)$ has depth
$|I_\text{curr}(a)|$, where an action node at each depth determines an
action of one agent in $I_\text{curr}(a)$. A leaf node of the tree at depth $|I_\text{curr}(a)|$
corresponds to a successor joint action $a_\text{next}\in \anext(a)$.

An \emph{action node} $\theta=(I_\text{tree},V_\text{tree},A_\text{tree})$
stores a partial assignment of new actions to the idle agents. Let $A_\text{tree}(\theta)$ denote the \emph{partial joint action} stored in $\theta$, i.e.,\ the new actions already assigned to some idle agents, and write $A^i_\text{tree}(\theta)$ for the action of agent $i$.
Let $I_\text{tree}(\theta)\subseteq I_\text{curr}(a)$ denote the set of idle agents whose actions are assigned in $A_\text{tree}(\theta)$ and $V_\text{tree}(\theta)$ the set of occupied vertices. Both are determined by $A_\text{tree}(\theta)$ and are kept in $\theta$ for convenience; two action nodes may share $I_\text{tree}$ and $V_\text{tree}$ but differ in $A_\text{tree}$.

The root node $\theta_0$ has $A_\text{tree}(\theta_0)=\emptyset$,
$I_\text{tree}(\theta_0)=\emptyset$ and
\begin{equation*}
\begin{aligned}
V_\text{tree}(\theta_0) ={}& \{v(a^i)\mid i\in I_\text{curr}(a)\}\\
&\cup \{p(a^j),\,v(a^j)\mid j\in I\setminus I_\text{curr}(a)\}.
\end{aligned}
\end{equation*}
For an action node $\theta$ at depth $d<|I_\text{curr}(a)|$, let $i_{d+1}$ be the next idle
agent whose action is to be assigned. Its destination is its current vertex
(a wait) or a neighbor of it (a move), but not an occupied vertex, i.e.,\
\begin{equation*}
\mathrm{Free}(\theta)
=\{v(a^{i_{d+1}})\}\cup \bigl(\mathrm{Neigh}(v(a^{i_{d+1}}))
\setminus V_\text{tree}(\theta)\bigr).
\end{equation*}
Choosing $u\in\mathrm{Free}(\theta)$ assigns agent $i_{d+1}$ the new action
\begin{equation*}
a^{i_{d+1}}_\text{next}=\bigl(v(a^{i_{d+1}}),\,u,\,\tau(a),\,t_v\bigr),
\end{equation*}
where $t_v=\tau(a)+\dur{i_{d+1}}{v(a^{i_{d+1}})}{u}$ for a move; for a wait,
$t_v$ is the next event, which is set once every idle agent is assigned.
Each choice gives a child that extends $\theta$ by this action:
\begin{equation*}
\begin{aligned}
\mathrm{Ch}(\theta)=\bigl\{\,
&\bigl(I_\text{tree}(\theta)\cup\{i_{d+1}\},\;
       V_\text{tree}(\theta)\cup\{u\},\\
&\;\;A_\text{tree}(\theta)\cup\{a^{i_{d+1}}_\text{next}\}\bigr)
 \;\big|\; u\in\mathrm{Free}(\theta)\,\bigr\}.
\end{aligned}
\end{equation*}

We label the edge $\theta\to\theta'$ by the appended action
$a^{i_{d+1}}_\text{next}$. A node stores the actions of all edges on its path
from $\theta_0$, so the partial joint action of any action node taken from the
frontier can be read directly from the node.

At a leaf $\theta$ (depth $|I_\text{curr}(a)|$), $A_\text{tree}(\theta)$ assigns a new action to every
idle agent; together with the actions kept by the busy agents, it determines a
successor $a_{\text{next}}\in\anext(a)$, and $\nextevent=\tau(a_{\text{next}})$.
Since every wait of $a_{\text{next}}$ ends at the next event, $\tau_{\text{next}}(a)$ is determined by
the move actions of $a_{\text{next}}$. Let $M$ be the set of these moves and let $d_{\min}$
be the minimum edge duration over all agents. Then
\[
\nextevent=
\begin{cases}
\min_{a^i\in M} t_v(a^i), & M\neq\emptyset,\\
\tau(a)+d_{\min}, & \text{otherwise},
\end{cases}
\]
and every wait action of $a_{\text{next}}$ has $t_v=\nextevent$.

Figure~\ref{fig:action_tree} (right) shows $AT(a)$ for two idle agents, so
$|I_\text{curr}(a)|=2$. At $\theta_0$, $A_\text{tree}(\theta_0)=\emptyset$ and
$V_\text{tree}(\theta_0)=\{A,D\}$. The first edge appends agent~1's move
$a^1_\text{next}=(A,B,\tau(a),\tau(a)+\dur{1}{A}{B})$, giving $\theta_1$ with
$A_\text{tree}(\theta_1)=\{a^1_\text{next}\}$, $I_\text{tree}(\theta_1)=\{1\}$
and $V_\text{tree}(\theta_1)=\{A,B,D\}$. The second edge appends agent~2's move
$a^2_\text{next}=(D,F,\tau(a),\tau(a)+\dur{2}{D}{F})$, giving $\theta_2$ with
$A_\text{tree}(\theta_2)=\{a^1_\text{next},a^2_\text{next}\}$,
$I_\text{tree}(\theta_2)=\{1,2\}$ and $V_\text{tree}(\theta_2)=\{A,B,D,F\}$.
Because $\theta_2$ is a leaf, it determines a successor joint action.

\subsubsection{Search tree}\label{method:preliminary:search tree}
Every joint action $a$ has its successors in
$\mathcal{A}_{\text{next}}(a)$, and each successor again has its own
successors. Starting from the initial joint action $a_0$, successors are generated recursively and the \emph{search tree} $\mathcal{S}$ is formed. 
A \emph{search tree} $\mathcal{S}$ represents a tree of joint actions. A tree node of $\mathcal{S}$ is a \emph{search node} $n$, which is a tuple $(a, n_{\mathrm{parent}},\mathcal{T},\lambda)$:
\begin{itemize}
    \item 
$a$ is a joint action represented by $n$. The root node $n_0$ carries the initial joint action
$a_0$. $n_{\text{next}}$ is a successor of $n$ if $a(n_{\text{next}})$ is a successor of $a(n)$. A search node whose joint action places every agent on its goal is a \emph{goal node}, and the path from $n_0$ to goal node is a solution (as defined in Sec.~\ref{lsrp-star-main:problem-def}). 
A \emph{search path} $P_{\mathrm{S}}(n)$ is a
unique sequence of search nodes $(n_0, n_1, \dots, n)$ from root to $n$. It represents a \emph{partial solution}. We write $g(n)=g(a(n))$ for the \emph{cost} of $n$. The root $n_0$ carries $a_0$ with $g(n_0)=0$.
    \item 
$n_{\mathrm{parent}}$ is the predecessor used to reconstruct a solution.

    \item 
$\mathcal{T}$ is the \emph{frontier} of the action tree $AT(a)$: the action nodes
of $AT(a)$ that have not been expanded yet. Action nodes in $\mathcal{T}$ are expanded in FIFO order. $\mathcal{T}$ is initialized to $\{\theta_0\}$ when $n$ is created.

    \item 
$\lambda$ is \emph{a rule based state} containing rule based properties to help complete actions of agents, which will be introduced in Sec.~\ref{rule-base-completion lsrp}.
\end{itemize}



For agent $i$ and vertex $u$, let $h^i(u)$ be a lower bound on the time
needed by $i$ to reach its goal from $u$. The heuristic of a search node $n$ is $h(n)=\sum_{i\in I}h^i(v(a^i(n)))$. Then $h(n)$ lower-bounds the remaining cost to
the goals from the current positions, so $f(n)=g(n)+h(n)$ is a lower bound
on the cost of any solution through $n$. We will use $f(n)$ to prune the search nodes that cannot improve on the incumbent.

Fig.~\ref{fig:action_tree} (left) illustrates these components.
Each search node owns an action tree, drawn as a small triangle.
The root $n_0$ carries $a_0$ and generates two successors, $n_j$ and $n_i$,
each of which owns a new action tree.
The gray successor $n_j$ is generated from a leaf, which assigns every idle
agent an action.
The orange successor $n_i$ is generated from a partial action node, whose
remaining idle agents' actions are assigned by the rule-based procedure
(Sec.~\ref{rule-base-completion lsrp}).
Action nodes of $AT(a(n_0))$ that have not been expanded remain in the
frontier $\mathcal T$.

\subsubsection{Dominance}

\begin{figure*}[tb]
  \centering
  \includegraphics[width=\linewidth]{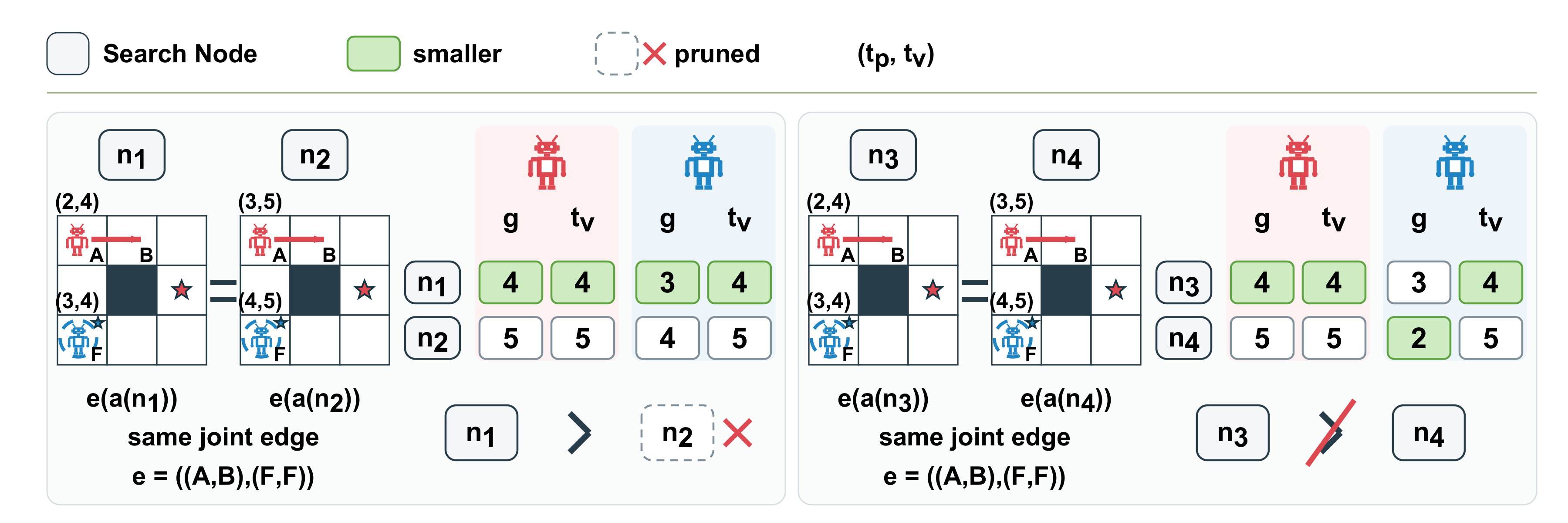}
  \caption{Dominance between search nodes with the same joint edge.
  Left: $n_1\dom n_2$; right: $n_3$ and $n_4$ are incomparable because
  an earlier goal arrival gives agent~2 a lower cost in $n_4$ despite
  its later action end time.}
  \label{fig:dominance}
\end{figure*}

Because agents finish actions at different times, two search nodes can reach the
same joint edge at different times. If one of them has an earlier arrival time and a lower cost
for every agent, the other can never lead to a cheaper solution.
\begin{definition}[Dominance, $\dom$]\label{def:dominance}
Let $n_1\neq n_2$ be two search nodes with $e(a(n_1))=e(a(n_2))$. Then $n_1$
\emph{dominates} $n_2$, written $n_1\dom n_2$, if
$g(a^i(n_1))\le g(a^i(n_2))$ and $t_v(a^i(n_1))\le t_v(a^i(n_2))$ for every $i\in I$.\footnote{This dominance is similar to the one in Loosely-Synchronized Search
(LSS)~\cite{ren2021loosely}, with one difference: here two search nodes are comparable only if they
have the same joint edge, whereas in LSS two joint states are comparable if they have the same end joint vertex. Comparing only the end joint vertex is not sufficient: an
agent moving along $(p,v)$ occupies both $p$ and $v$, so two joint states with the same end joint vertex but different $p$ may block different vertices, and the pruned joint state may be the only one that leads to an optimal solution. See Appendix~\ref{app:lss} for an example.
Besides, we checked the original implementation of LSS~\cite{ren2021loosely}, which is the same as the dominance rule (i.e., decide whether two states are comparable based on the joint edge, rather than the ending joint vertex) reported in this paper, and is correct.
The reported numerical results in LSS~\cite{ren2021loosely} are still correct.
However, the writing of the dominance rule in the LSS paper~\cite{ren2021loosely} is wrong, and should be replaced by Definition~\ref{def:dominance}.
}
\end{definition}

Fig.~\ref{fig:dominance} illustrates Definition~\ref{def:dominance} with asynchronous actions.
Agent~1 needs two time units to move from $A$ to $B$, while agent~2 is at its
goal $F$ and waits for one time unit. In each panel the two search nodes have
the same joint edge $e=((A,B),(F,F))$ and differ only in when the actions take
place. In $n_1$, agent~1 moves during $[2,4]$. 
Agent~2 arrives at its goal $F$
at time $3$ and waits during $[3,4]$; its wait ends at the next event, which
is the arrival of agent~1 at time $4$. Because agent~2 has reached its goal at
time $3$, its cost is frozen at $g=3$: the wait only increases its arrival
time to $t_v=4$, but not $g$. 
In $n_2$, the same actions happen one time unit
later: agent~1 moves during $[3,5]$, and agent~2 reaches $F$ at time $4$ and
waits during $[4,5]$, giving $g=4$ and $t_v=5$. Every entry of $n_1$ is
smaller than the corresponding entry of $n_2$, hence $n_1\succ n_2$ and $n_2$
is pruned. 
In the right panel, $n_4$ differs from $n_3$ in that agent~2
reached its goal earlier, at time $2$ instead of $3$, and then waited until
time $5$; thus its $g$ is frozen at $2$, smaller than $g=3$ in $n_3$, while its
$t_v=5$ is larger than $t_v=4$ in $n_3$. Agent~1 also arrives later in $n_4$.
Neither node is smaller in every entry, so neither dominates the other and both are kept. 
Their sums of cost are equal, $g(n_3)=g(n_4)=7$, which does not help with pruning, and $g$ and $t_v$ needs to be compared for every agent.

\subsection{Algorithm Overview}\label{sec:overview}

\ours is a depth-first, anytime search over the joint actions of all agents.
It has two parts: (i) A search algorithm (Sec. \ref{method:high-level search}) expands one successor per iteration toward goal, so that a first solution is reached quickly, and then keeps improving it until it reaches an optimal solution or the time budget runs out. (ii) A rule-based procedure makes a successor joint action completed.

The search algorithm provides the anytime behavior. Starting from the initial
joint action, it repeatedly pops a search node and expands one action node of its
action tree to assign actions of a subset of idle agents. Then it asks the rule-based procedure to complete the actions of the
remaining agents. The resulting successor is inserted only if it is not
dominated, and a goal successor updates the incumbent. Lazy expansion, which
generates a single successor per iteration, lets a first solution be reached
quickly. The search then keeps refining the incumbent through three
mechanisms: (i) dominance discards a search node that reaches the same joint edge with no larger cost and no later arrival time for every agent; 
(ii) \emph{re-initialization} (re-init) periodically restarts from the root to begin a new search path instead of deepening the current one; and 
(iii) an admissible $f$-bound discards nodes that cannot improve
the incumbent.

The rule-based procedure completes the actions of the agents left unassigned by the action tree.
It keeps a priority for each agent and an action cache of deferred actions.
It resolves conflicts by pushing other agents out of the way (\textsc{AsyPush}),
and tries the candidate vertices of each agent in the order given by \textsc{Order}:
\textsc{Swap} lets two agents exchange positions in a corridor,
and \textsc{Flip} breaks distance ties by estimated push delays.

Together, the two parts
let \ours explore the joint-action space efficiently while retaining the
termination, completeness, and optimality guarantees established in
Sec.~\ref{lsrp-star-main:analysis}.

\subsection{LSRP*: Search Algorithm}\label{method:high-level search}

\ours{} is a depth-first, anytime search over a search tree (Sec.~\ref{method:preliminary:search tree}). It contains three parts: (i) Main search loop: \ours maintains the cheapest solution found so far and keeps improving it until the time budget is depleted. (ii) Successor expansion: \textsc{LSRPExpd} expands a search node $n$ into one successor. It expands an action node from action tree (Sec.~\ref{method:preliminary:action tree}) of $n$ to assign the actions of a few idle agents, and a rule-based procedure completes the
rest. (iii) Dominance: the newly generated successor is tested against 
dominance through \textsc{AddNew}.

\subsubsection{Main Search Loop}
The search has three key features.
First, the search nodes are kept in a stack $Open$, so the search dives to a goal node quickly and then keeps improving it, which makes \ours
anytime. Second, \ours{} is depth-first, but search nodes are expanded lazily: standard DFS generates all successors of a node at once, but \ours generates a single successor for each expansion. The search node stays in $Open$ until it cannot be expanded anymore. Third, the search is re-initialized periodically: after a number of iterations, it expands the root $n_0$ instead of the top of $Open$, so that a new search path with a different first joint action is explored. This avoids deepening the current path when it yields little improvement. Lazy expansion is what makes the first solution fast; re-init is what makes the subsequent improvement effective. Both are discussed in
Remarks~\ref{rem:lazy} and \ref{rem:reinit} after the algorithm.

\begin{algorithm}[tb]
\small
\caption{\textsc{LSRP*}}\label{alg:lsrp*}
\begin{algorithmic}[1]
\State $(c, c_{\max}) \gets (0, 0)$ \label{alg:lsrp*:init_c}
\State $(Open, \mathcal{F}, n^*) \gets (\{n_0\}, \emptyset, \emptyset)$ \label{alg:lsrp*:init_vars}
\While{$Open \neq \emptyset \land \mathbf{not}\ \text{timeout}$} \label{alg:lsrp*:while}
    \State $n \gets Open.\text{pop}()$; $c \gets c + 1$ \label{alg:lsrp*:pop}
    \If{$n^* \neq \emptyset \land c = c_{\max}$} \Comment{re-init} \label{alg:lsrp*:reinit_check}
        \State $Open.\text{push}(n)$ \label{alg:lsrp*:push_n}
        \State $(n, c) \gets (n_0, 0)$ \Comment{change to the root instead} \label{alg:lsrp*:pop_root}
    \EndIf \label{alg:lsrp*:reinit_endif}
    \If{$\mathcal{T}(n) = \emptyset$} \textbf{continue} \Comment{action tree exhausted} \label{alg:lsrp*:check_exhausted}
    \EndIf \label{alg:lsrp*:exhausted_endif}
    \If{$n^* \neq \emptyset \land f(n) \ge g(n^*)$} \textbf{continue} \Comment{f-prune test} \label{alg:lsrp*:check_fprune}
    \EndIf \label{alg:lsrp*:fprune_endif}
    \State $(\partial n, n_{\text{next}}) \gets \textsc{LSRPExpd}(n)$ \Comment{$\partial n$: residual of $n$} \label{alg:lsrp*:expd}
    \If{$\mathcal{T}(\partial n) \neq \emptyset$} \label{alg:lsrp*:check_residual}
        \State $Open.\text{push}(\partial n)$ \Comment{reuse the residual} \label{alg:lsrp*:push_residual}
    \EndIf \label{alg:lsrp*:residual_endif}
    \State $(Open, \mathcal{F}) \gets \textsc{AddNew}(n_{\text{next}}, Open, \mathcal{F})$ \label{alg:lsrp*:addnew}
    \If{$\textbf{v}(a(n_{\text{next}})) = \textbf{v}_g \land (n^* = \emptyset \lor g(n_{\text{next}}) < g(n^*))$} \label{alg:lsrp*:check_goal}
        \State $(c_{\max}, c) \gets (c, 0)$ \label{alg:lsrp*:update_c}
        \State $n^* \gets n_{\text{next}}$ \label{alg:lsrp*:update_nstar}
    \EndIf \label{alg:lsrp*:goal_endif}
\EndWhile \label{alg:lsrp*:endwhile}
\State \Return $\textsc{PostProcess}(n^*)$ \Comment{return the best solution found} \label{alg:lsrp*:return}
\end{algorithmic}
\end{algorithm}

The search starts from the root $n_0$, with no incumbent  (lines~\ref{alg:lsrp*:init_c}--\ref{alg:lsrp*:init_vars}).
Each iteration pops a node $n$ and increments the iteration counter $c$
(line~\ref{alg:lsrp*:pop}). If an incumbent already exists and the search has spent
as many iterations as the previous improvement needed ($c = c_{\max}$), the popped
node is put back into $Open$ and the root is expanded instead
(lines~\ref{alg:lsrp*:reinit_check}--\ref{alg:lsrp*:pop_root}); this is denoted the re-init process, discussed in Remark~\ref{rem:reinit}. Two tests may then prune $n$: its action tree frontier $\mathcal{T}$ is exhausted, in which case $n$ has no successor left (line~\ref{alg:lsrp*:check_exhausted}); or its $f$-value may already reach the cost
of the incumbent, in which case no successor of $n$ can improve it (line~\ref{alg:lsrp*:check_fprune}). Otherwise \textsc{LSRPExpd} expands $n$ and returns both the successor $n_{\text{next}}$ and the residual $\partial n$, which is the same node with a smaller frontier (line~\ref{alg:lsrp*:expd}). The
residual is pushed back to $Open$ whenever its frontier is still non-empty
(lines~\ref{alg:lsrp*:check_residual}--\ref{alg:lsrp*:push_residual}), so $n$ can be
expanded again later along a different joint action. The successor $n_{\text{next}}$ is given to \textsc{AddNew} (line~\ref{alg:lsrp*:addnew}), which inserts it into $Open$ only if it is not pruned by dominance. Finally, if the successor is a goal node that is cheaper than the incumbent, the incumbent is replaced and $c_{\max}$ is reset to the number of iterations this improvement took (lines~\ref{alg:lsrp*:check_goal}--\ref{alg:lsrp*:update_nstar}). When $Open$ empties or the time limit is reached, the best solution found so far is returned  through $\textsc{PostProcess}$ (line~\ref{alg:lsrp*:return}). 

\begin{figure*}[tb]
  \centering
  \includegraphics[width=0.9\linewidth]{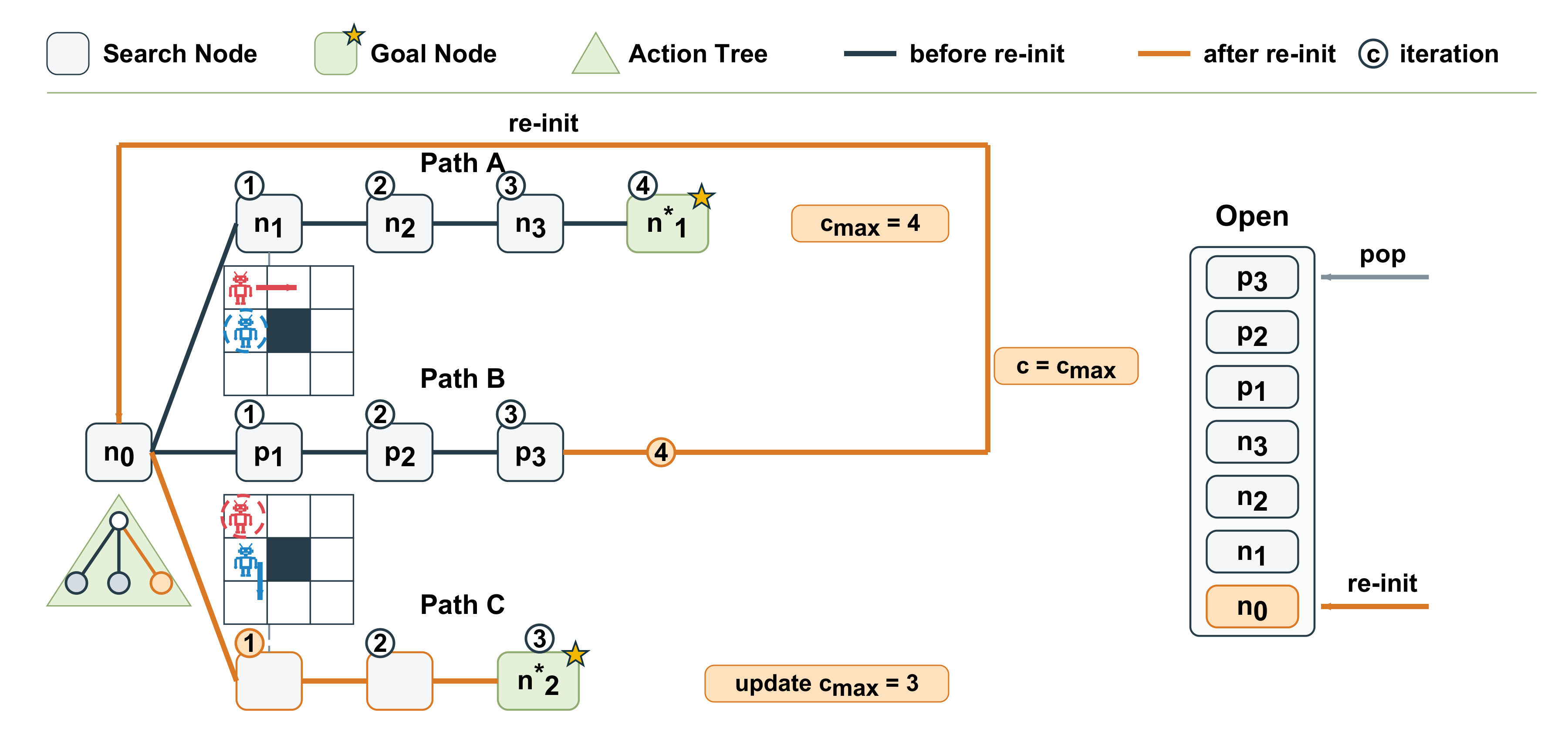}
  \caption{Illustration of re-init.
  Circled numbers count iterations $c$: Path~A reaches the first goal
  $n^*_1$ at $c=4$, so $c_{\max}=4$.
  When Path~B reaches no goal by $c=c_{\max}$, re-init (orange) expands
  $n_0$ instead of the next popped node, and Path~C reaches a cheaper
  goal $n^*_2$.}
  \label{fig:reinit}
\end{figure*}

\begin{remark}[Lazy expansion]\label{rem:lazy}

A search node $n$ has $|\mathcal{A}_{\text{next}}(a(n))|$ successors, which grows exponentially with the number of agents, so generating them all at once (as standard DFS) wastes time on enumeration. Instead, \ours{} uses \emph{lazy expansion}, generating one successor per iteration and keeping the parent in $Open$, driving one search path deeper toward the goal. Thus a first solution is reached fast and the saved time is spent improving the incumbent.
\end{remark}

\begin{remark}[Re-init]\label{rem:reinit}
Once an incumbent exists, every $c_{\max}$ iterations, \ours{} expands the
root $n_0$ instead of the current node $n$. Since action nodes are expanded in
FIFO order, this starts a new search path from an unexplored first joint
action, which may reach a solution cheaper than $g(n^*)$. We set $c_{\max}$ to
the number of iterations that the last improvement took, so a search path that
fails to improve within this budget is treated as \emph{unpromising}. Re-init
only changes which node is popped next: it prunes no search node and never
changes the incumbent, so completeness (Theorem~\ref{thm:complete}) is
unaffected.

Fig.~\ref{fig:reinit} gives an example. Path~A reaches the first goal $n^*_1$
at $c=4$, which sets $c_{\max}=4$ and resets $c$. Path~B then finds no goal,
and when $c$ reaches $c_{\max}$, the popped node is put back into \textit{Open}
and $n_0$ is expanded instead. Its next action node yields a different first joint action, and Path~C reaches a cheaper goal $n^*_2$ within
three iterations, which sets $c_{\max}$ to 3. The stack on the right shows
\textit{Open} at the moment of re-init: all nodes $n_0,\dots,p_3$ are still
stored, and only the next node to pop changes from $p_3$ to $n_0$.
\end{remark}

\begin{algorithm}[tb]
\small
\caption{Expansion: \textsc{LSRPExpd}}\label{alg:lsrpexpd}
\begin{algorithmic}[1]
\Procedure{\textsc{LSRPExpd}}{$n$} \label{alg:lsrpexpd:proc}
   \State $\theta \gets \text{Pick}(\mathcal{T}(n))$\label{alg:lsrpexpd:pick}
    \State $\partial n \gets n$ \textbf{with} $\mathcal{T}(\partial n) = (\mathcal{T}(n) \setminus \{\theta\}) \cup \text{Ch}(\theta)$
    \label{alg:lsrpexpd:residual}
    \State $\partial a_{next} \gets \emptyset$ \label{alg:lsrpexpd:init_A}
    \For{$i \notin I_{\text{curr}}(a(n))$} \Comment{Busy agents keep their action} \label{alg:lsrpexpd:for_intransit}
    \State $\partial a_{next}^i \gets a^i(n)$ \label{alg:lsrpexpd:assign_intransit}
    \EndFor \label{alg:lsrpexpd:endfor_intransit}
    \For{$i \in I_{\text{tree}}(\theta)$} \Comment{actions already assigned by action tree} \label{alg:lsrpexpd:for_fixed}
    \State $\partial a_{next}^i \gets A_{\text{tree}}^i(\theta)$ \label{alg:lsrpexpd:assign_fixed}
\EndFor \label{alg:lsrpexpd:endfor_fixed}
    \State $(\lambda_{\text{next}}, a_{\text{next}}) \gets \textsc{LSRPStep}(n,\, \partial a_{next}, \theta \text{ is $\theta_0$})$
       \label{alg:lsrpexpd:lsrpstep}
    \State $n_{\text{next}} \gets (a_{\text{next}}, n_{\text{parent}}=n, \mathcal{T}=\{\theta_0\}, \lambda_{\text{next}})$ \label{alg:lsrpexpd:searchnode}
    \State \Return $(\partial n, n_{\text{next}})$ \label{alg:lsrpexpd:return}
\EndProcedure \label{alg:lsrpexpd:endproc}
\end{algorithmic}
\end{algorithm}

\begin{algorithm}[tb]
\small
\caption{\textsc{AddNew}}\label{alg:addnew}
\begin{algorithmic}[1]
\Procedure{\textsc{AddNew}}{$n, Open, \mathcal{F}$} \label{alg:addnew:proc}
    \If{exists $m\in \mathcal{F}(e(a(n)))$ and $m\dom n$} \label{alg:addnew:check_dominated}
        \State \Return $(Open, \mathcal{F})$ \Comment{prune dominated candidate} \label{alg:addnew:return_discard}
    \EndIf \label{alg:addnew:endif_dominated}
    \State $D \gets \{m \in \mathcal{F}(e(a(n))) \mid n \text{ \dom } m\}$\label{alg:addnew:find_dominated}
    \State $Open \gets Open \setminus D$ \label{alg:addnew:remove_open}
    \State $\mathcal{F}(e(a(n))) \gets \mathcal{F}(e(a(n))) \setminus D$ \label{alg:addnew:remove_f}
    \State $Open.\text{push}(n)$ \label{alg:addnew:push_open}
    \State $\mathcal{F}(e(a(n))).\text{add}(n)$ \label{alg:addnew:add_f}
    \State \Return $(Open, \mathcal{F})$ \label{alg:addnew:return}
\EndProcedure \label{alg:addnew:endproc}
\end{algorithmic}
\end{algorithm}

\subsubsection{Successor Expansion}\label{method:high-level-expansion:lsrpexpd}

\textsc{LSRPExpd} (Algorithm \ref{alg:lsrpexpd}) expands a search node $n$ by expanding one action node $\theta$ into one successor: (i) Given a search node $n$, the actions of the busy agents ($I\setminus I_{\text{curr}}(a(n))$) are kept from previous joint action. (ii) The actions of agents in $I_{\text{tree}}(\theta)$, a subset of idle agents $I_{\text{curr}}(a(n))$, are read from $A_\text{tree}(\theta)$ (Sec. \ref{method:preliminary:action tree}). (iii) The actions of the rest of the agents ($I_{\text{curr}}(a(n)) \setminus I_{\text{tree}}(\theta)$) are completed by a rule-based procedure (Sec.~\ref{rule-base-completion lsrp}). 

An action node $\theta$ is picked from the frontier $\mathcal{T}(n)$ (line~\ref{alg:lsrpexpd:pick}). The residual node $\partial n$ is the same search node as $n$ in which $\theta$ has been replaced by its children $\mathrm{Ch}(\theta)$ (line~\ref{alg:lsrpexpd:residual}). The action nodes $\mathrm{Ch}(\theta)$ are kept for later expansions of $n$. The partial joint action $\partial a_{next}$ is
then assembled in three steps: (i) Busy agents simply keep their actions (lines~\ref{alg:lsrpexpd:for_intransit}--\ref{alg:lsrpexpd:endfor_intransit}). (ii) Agents in $I_{\text{tree}}(\theta)$ take the actions stored in $A_\text{tree}(\theta)$ (lines~\ref{alg:lsrpexpd:for_fixed}--\ref{alg:lsrpexpd:endfor_fixed}). (iii) The remaining agents of $I_{\text{curr}}(a(n))$ have their actions completed by a rule-based procedure \textsc{LSRPStep}, which returns the completed joint action $a_{\text{next}}$ together with the updated rule-based state $\lambda_{\text{next}}$ (line~\ref{alg:lsrpexpd:lsrpstep}). A new
search node is created from that joint action, with $n$ as its parent and a new action tree consisting of the root action node $\theta_0$ alone
(line~\ref{alg:lsrpexpd:searchnode}). The procedure returns the residual
and the successor (line~\ref{alg:lsrpexpd:return}).

\subsubsection{Dominance}\label{sec:dominance}
\textsc{AddNew} (Algorithm~\ref{alg:addnew}) is the only place where a search node enters $Open$ and is tested against dominance (Definition~\ref{def:dominance}). It keeps a \emph{dominance
index} $\mathcal{F}$ of the search nodes generated so far, grouped by their joint edge $e(a(n))$: a candidate $n$ is pruned if some search node in $\mathcal{F}(e(a(n)))$ dominates it, and otherwise it is inserted into
$Open$ and $\mathcal{F}$. Similarly, the nodes that $n$ dominates are removed from $Open$ and $\mathcal{F}$. $Open$ thus contains only non-dominated search nodes.

\subsection{Rule-based Completion: LSRP}\label{rule-base-completion lsrp}

\begin{algorithm}[tb]
\small
\caption{Rule-based collective step: \textsc{LSRPStep}}\label{alg:lsrpstep}
\begin{algorithmic}[1]
\Procedure{\textsc{LSRPStep}}{$n, \partial a_{next}, \text{isRoot}$} \label{alg:lsrpstep:proc}
    \State $\epsilon,\Phi \gets \lambda(n)$
    \State $\epsilon_{\text{next}} \gets \epsilon$ \label{alg:lsrpstep:copy_eps}
    \For{$i \in I$ \textbf{with} $v^i = v_g^i$} \Comment{reset when reaching goal} \label{alg:lsrpstep:for_goal}
        \State $\epsilon_{\text{next}}^i \gets 0$ \label{alg:lsrpstep:reset_eps}
    \EndFor \label{alg:lsrpstep:endfor_goal}
    \State $\Phi_{\text{next}} \gets \emptyset$ \Comment{deferred actions for the child} \label{alg:lsrpstep:init_phi}
    \If{$\text{isRoot}$} \label{alg:lsrpstep:inherited_cache_root_action_node}
        \For{\textbf{each} agent $i$ with a pending action $\Phi^i$} \label{alg:lsrpstep:for_phi}
            \If{$t_p(\Phi^i) = \tau$} \label{alg:lsrpstep:if_time}
                \State $\partial a_{next}^i \gets \Phi^i$ \Comment{its time has come: move now} \label{alg:lsrpstep:move_now}
            \Else \label{alg:lsrpstep:else_time}
                \State $\partial a_{next}^i \gets \text{act}(v^i, v^i, \tau,\tau_{\text{next}})$ \label{alg:lsrpstep:wait}
                \State $\Phi_{\text{next}}^i \gets \Phi^i$ \Comment{still waiting: carry forward} \label{alg:lsrpstep:carry_forward}
            \EndIf \label{alg:lsrpstep:endif_time}
        \EndFor \label{alg:lsrpstep:endfor_phi}
    \EndIf
     \For{$i \in I_{\text{curr}}(a(n))$ in descend $\epsilon_{\text{next}}$ \textbf{with} $\partial a_{next}^i = \emptyset$} \label{alg:lsrpstep:for_icurr}
        \State $(\_, \Phi_{\text{next}}, \partial a_{next}) \gets \textsc{AsyPush}(i, a(n), \partial a_{next}, \Phi_{\text{next}})$ \label{alg:lsrpstep:onestep}
    \EndFor
    \State $\lambda_{\text{next}}\gets (\epsilon_{\text{next}},\Phi_{\text{next}})$
    \State \Return $(\lambda_{\text{next}},  a_{next})$ \Comment{$\partial a_{\text{next}}$ is complete} \label{alg:lsrpstep:return}
\EndProcedure \label{alg:lsrpstep:endproc}
\end{algorithmic}
\end{algorithm}
\begin{algorithm}[tb]
\small
\caption{Single rule-based step: \textsc{AsyPush}}\label{alg:asy_push}
\begin{algorithmic}[1]
\Procedure{\textsc{AsyPush}}{$i, a, \partial a_{next}, \Phi$} \label{alg:asy_push:proc}
    \State $P \gets \{\, k \in I \;|\; \partial a_{next}^k \neq \emptyset \,\}$ \label{alg:asy_push:P}
    \State $U \gets \{\, p^k_{next} ,\; v^k_{next} \;|\; k \in P \,\}$ \label{alg:asy_push:U}
    \State $H \gets \{\, v^k \mapsto k \;|\; k \in I \setminus (P \cup \{i\}) \,\}$ \label{alg:asy_push:H}
    \State $C \gets \textsc{Order}(i, \partial a_{next}, H)$ \Comment{Algorithm~\ref{alg:order}} \label{alg:asy_push:C}
    \For{$v \in C$} \label{alg:asy_push:for_v}
        \If{$v \in U$} \textbf{continue} \label{alg:asy_push:check_U}
        \EndIf
        \If{$v \in H$} \label{alg:asy_push:if_dom}
            \State $j \gets H[v]$ \label{alg:asy_push:get_j}
            \State $\partial a_{next}^i \gets \text{act}(v^i, v^i, \tau, \tau_{\text{next}})$ \Comment{$i$ waits at $v^i$} \label{alg:asy_push:hold_vi}
            \State $(t^{\text{rel}}_j, \Phi, \partial a_{next}) \gets \textsc{AsyPush}(j, a, \partial a_{next}, \Phi)$ \label{alg:asy_push:recursive}
            \If{$t^{\text{rel}}_j = \bot$} $\partial a_{next}^i \gets \emptyset$; \textbf{continue} \label{alg:asy_push:check_bot}
            \EndIf
            \State $\Phi^i \gets \text{act}(v^i, v, t^{\text{rel}}_j, \bot)$ \Comment{deferred move of $i$} \label{alg:asy_push:update_phi}
            \State $t^{\text{rel}}_i \gets t^{\text{rel}}_j + D(i, v^i, v)$ \label{alg:asy_push:t_push}
        \Else \Comment{$v$ is free --- move now} \label{alg:asy_push:else_free}
            \State $\partial a_{next}^i \gets \text{act}(v^i, v, \tau, \tau_{\text{next}})$ \label{alg:asy_push:assign_v}
            \State $t^{\text{rel}}_i \gets t_v(\partial a^i_{next})$ \Comment{$i$ arrives at $v$} \label{alg:asy_push:t_free}
        \EndIf
        \State \Return $(t^{\text{rel}}_i, \Phi, \partial a_{next})$ \label{alg:asy_push:return}
    \EndFor \label{alg:asy_push:endfor_v}
    \State $\partial a_{next}^i \gets \text{act}(v^i, v^i, \tau, \tau_{\text{next}})$ \Comment{nothing worked: wait} \label{alg:asy_push:wait}
    \State \Return $(\bot, \Phi, \partial a_{next})$ \Comment{no move; failure} \label{alg:asy_push:return_bot}
\EndProcedure \label{alg:asy_push:endproc}
\end{algorithmic}
\end{algorithm}
Expanding an action node $\theta$ assigns the actions of the agents in
$I_{\text{tree}}(\theta)$ only.
To obtain a successor without waiting for enumerating every idle agent's action in the action tree to reach a leaf, LSRP* applies a rule-based procedure to complete the actions of the remaining unplanned agents. \textsc{LSRPStep}
(Algorithm~\ref{alg:lsrpstep}) visits the
unplanned agents in a rule-based order and calls \textsc{AsyPush}
(Algorithm~\ref{alg:asy_push}) to assign each one a single action. The
completed joint action corresponds to a leaf of $AT(a)$.
\begin{figure*}[tb]
  \centering
  \includegraphics[width=\linewidth]{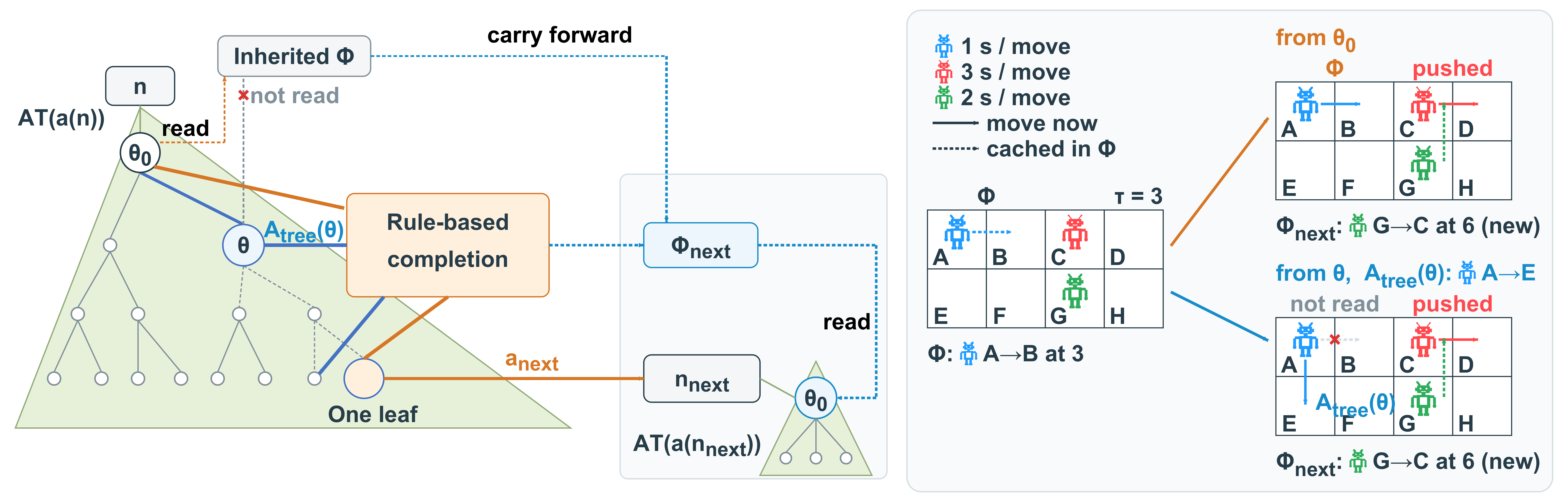}
  \caption{Rule-based completion and the action cache $\Phi$.
Left: completions from the root $\theta_0$ (orange) and a non-root
action node $\theta$ (blue); only the root call reads the inherited
$\Phi$, and only the orange successor is shown.
Right: an example at $\tau=3$.}
  \label{fig:phi_lifecycle}
\end{figure*}

Each search node $n$ carries a rule-based state
$\lambda(n)=(\epsilon,\Phi)$.
The priority vector $\epsilon$ determines the processing order of
unplanned idle agents, and the action cache $\Phi$ stores deferred
moves, where $t_p(\Phi^i)$ is the scheduled start time of agent $i$'s
deferred move.
The cache is needed when an agent must wait for another agent to
release its intended destination.
Since a joint action contains only one action per agent, the current
wait and the deferred move cannot both be represented in
$a_{\mathrm{next}}^i$.
Moreover, the destination may be released later than the next event
$\tau_{\mathrm{next}}$, because other agents may complete their
actions earlier.
The current wait therefore ends at $\tau_{\mathrm{next}}$, while
$\Phi$ preserves the deferred move across intermediate events until its scheduled start time.

For an idle agent $i$ that will start a new action at event $\tau$,
with next event $\tau_{\text{next}}$, we write
$\mathrm{act}(v^i, u^i, \tau,\tau_{\text{next}})$ as a function that
rebuilds an action, where $v^i$ is the vertex that $i$ currently
occupies and $u^i$ is the next vertex it is going to occupy:
\begin{equation}
\text{act}(v^i, u^i, \tau, \tau_{\text{next}}) =
\begin{cases}
(v^i, u^i, \tau, \tau + D(i, v^i, u^i)) & \text{move},\\
(v^i, v^i, \tau, \tau_{\text{next}}) & \text{wait}
\end{cases}
\end{equation}
A deferred move is written $\text{act}(v^i, u^i, t_p, \bot)$: it starts
at $t_p$, and its end time is set only when it is assigned.

The state $\lambda$ is produced by the rule-based completion that
generates $n$ from its parent and is stored in $n$.
When $n$ is expanded, we call its stored cache the \emph{inherited}
cache, since it was created during the parent's expansion.
Storing $\lambda$ in each search node therefore preserves priorities
and deferred moves along the search path.
Figure~\ref{fig:phi_lifecycle} (left) shows two separate rule-based
completion calls within the same action tree.
The orange route starts at the root $\theta_0$, where no idle-agent
action is assigned by the tree.
Inherited cache $\Phi$ is processed before the remaining agents are
assigned actions: entries due at current event $\tau$ are used to assign moves and removed from the output cache, while entries that are not yet due are carried forward.
The completed joint action corresponds to the orange leaf and
generates the successor $n_{\mathrm{next}}$.
Its output cache $\Phi_{\mathrm{next}}$ contains any carried-forward
entries and any new deferred moves generated during this completion.
The successor's cache can subsequently be read and processed at
the root of its own action tree.

The blue route instead starts at a non-root action node $\theta$.
This call keeps the actions in $A_\text{tree}(\theta)$ assigned to the agents in $I_{\text{tree}}(\theta)$ and completes
the remaining actions of unplanned idle agents in $I_{\text{curr}}(a(n))\setminus I_{\text{tree}}(\theta)$ without reading or carrying forward inherited cache $\Phi$.
This call may generate new deferred moves in its own output
cache, and it will be read in the newly generated action tree when expanding its root action node.
The two routes represent alternative completion attempts at the same
event. For clarity, only the orange route's successor and cache transfer are shown in figure~\ref{fig:phi_lifecycle} (left).

Given $n$, a partial joint action $\partial a_{\text{next}}$, and a flag
$\text{isRoot}$ indicating whether the current action node is the root
$\theta_0$,
\textsc{LSRPStep} copies the priorities from $\lambda(n)$ into
$\epsilon_{\text{next}}$ and resets the priority of every agent at its goal to
zero, so that agents still far from their goal are planned first
(lines~\ref{alg:lsrpstep:copy_eps}--\ref{alg:lsrpstep:endfor_goal}). If the action node is the root, it then
processes the deferred actions in $\Phi$
(lines~\ref{alg:lsrpstep:init_phi}--\ref{alg:lsrpstep:endfor_phi}): an action
due at the current event time $\tau$ is assigned to its agent
(line~\ref{alg:lsrpstep:move_now}); otherwise the agent waits over
$[\tau,\tau_{\text{next}}]$ and carries its deferred action forward in
$\Phi_{\text{next}}$
(lines~\ref{alg:lsrpstep:wait}--\ref{alg:lsrpstep:carry_forward}). Finally, the
remaining unplanned agents in $I_{\text{curr}}(a(n))$ are processed in
descending order of $\epsilon_{\text{next}}$; for each, \textsc{AsyPush}
assigns an action, recursively assigns actions to the agents blocking it, and
updates $\Phi_{\text{next}}$ as needed
(lines~\ref{alg:lsrpstep:for_icurr}--\ref{alg:lsrpstep:onestep}).
\textsc{LSRPStep} returns the completed joint action $a_{\text{next}}$ and the
updated state $\lambda_{\text{next}}=(\epsilon_{\text{next}},\Phi_{\text{next}})$
(line~\ref{alg:lsrpstep:return}).

Figure~\ref{fig:phi_lifecycle} (right) gives an example at event
$\tau = 3$ with three agents whose moves take $1$\,s (blue), $3$\,s
(red) and $2$\,s (green).
The inherited cache holds one entry, the blue agent's move from $A$
to $B$ with $t_p = 3$.
In the orange route from $\theta_0$, this entry is due, so the blue
agent moves to $B$ at once and the entry is removed.
\textsc{AsyPush} then processes the green agent: its preferred vertex
$C$ is occupied by the unplanned red agent, which is pushed from $C$
to $D$ and releases $C$ at $3 + 3 = 6$.
The green agent therefore waits and stores the new entry
$\text{act}(G, C, 6, \bot)$ in $\Phi_{\mathrm{next}}$.
The next event is $\tau_{\mathrm{next}} = 4$, when the blue agent
reaches $B$, which is earlier than the release of $C$.
The green agent's wait thus ends at $4$; when $n_{\mathrm{next}}$ is
expanded from its root, the entry is not yet due and is carried
forward, until it becomes due at event $6$.
In the blue route, $A_{\mathrm{tree}}(\theta)$ assigns the blue agent
the action from $A$ to $E$.
The inherited entry from $A$ to $B$ is not read and is dropped, while
the red and green agents are completed as in the orange route.
Both routes therefore output the same new entry from $G$ to $C$ at
$6$, but only the orange route consumes the inherited entry.

\begin{algorithm}[tb]
\small
\caption{Candidate ordering: \textsc{Order}}\label{alg:order}
\begin{algorithmic}[1]
\Procedure{\textsc{Order}}{$i, \partial a_{next}, H$} \label{alg:order:proc}
    \State $C \gets \text{Neigh}(v^i)$ \label{alg:order:C}
    \If{$i$ is not being pushed} $C \gets C \cup \{v^i\}$ \Comment{a pushed agent must leave $v^i$} \label{alg:order:wait}
    \EndIf
    \State sort $C$ by $D(i, v^i, v) + h^i(v)$ ascending \label{alg:order:sort}
    \If{\textsc{Swap}$^\dagger$ is enabled $\wedge$ \textsc{Swap} is required for $i$} \label{alg:order:swap}
        \State reverse $C$ \Comment{$^\dagger$swap rule of LSRP, Remark~\ref{rem:swap}} \label{alg:order:reverse}
    \ElsIf{$n^* \neq \emptyset$} \Comment{\textsc{Flip}, after the first solution} \label{alg:order:if_flip}
        \For{$v \in C$} \label{alg:order:for_delay}
            \State $\delta(v) \gets \textsc{SimPush}(H[v], \partial a_{next})$ \textbf{if} $v \in H$ \textbf{else} $0$ \label{alg:order:delay}
        \EndFor
        \State stable-sort $C$ by $\delta(v)$ ascending within each tie of $D(i, v^i, v) + h^i(v)$ \label{alg:order:tie}
    \EndIf
    \State \Return $C$ \label{alg:order:return}
\EndProcedure
\end{algorithmic}
\end{algorithm}

\subsubsection{Single-agent Step}\label{method:asypush}

\begin{figure*}[tb]
  \centering
  \includegraphics[width=\linewidth]{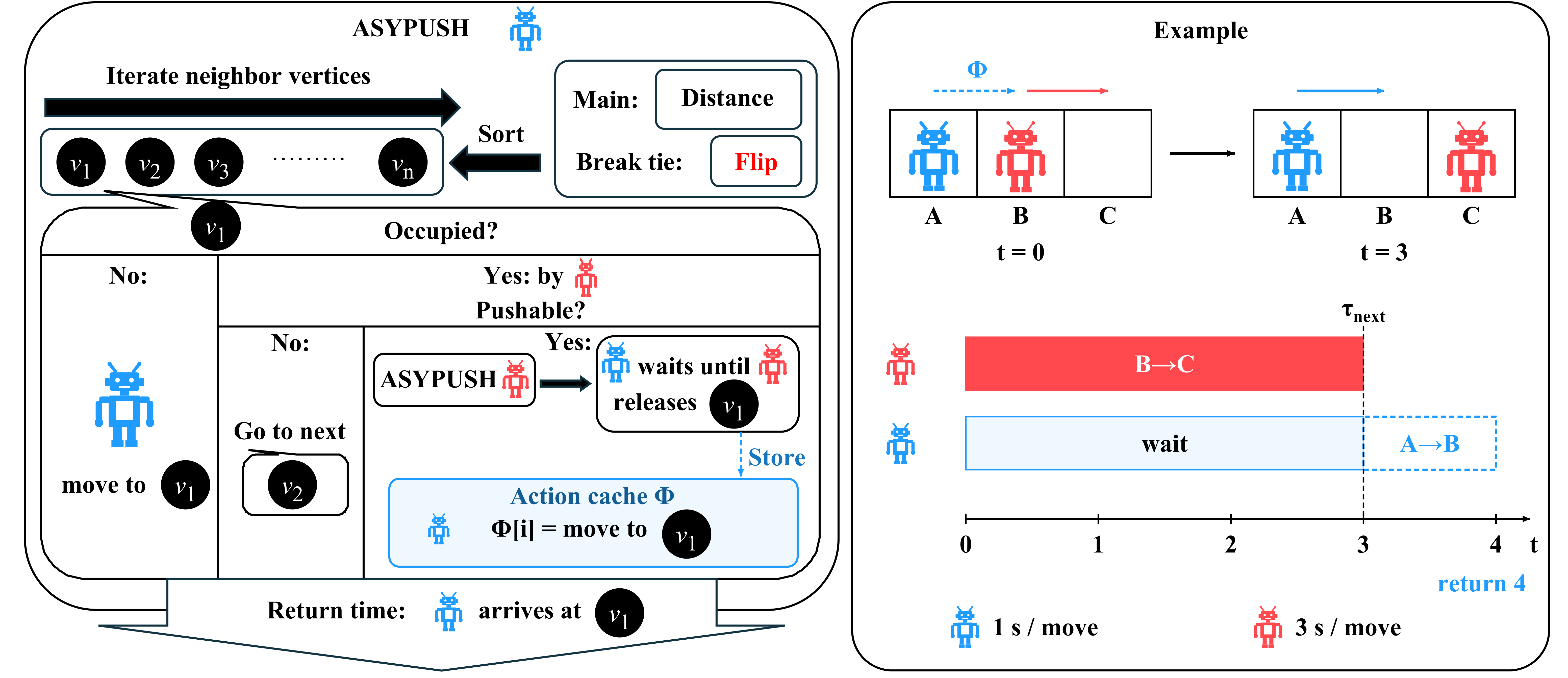}
  \caption{Single-agent action assignment by \textsc{AsyPush}.
Left: a successful recursive push makes the processing agent wait and
store its deferred move in $\Phi$.
Right: an example in a corridor.}

  \label{fig:asy_push}
\end{figure*}

\textsc{AsyPush} (Algorithm~\ref{alg:asy_push}) assigns an action
to a single agent through recursive pushing.
Figure~\ref{fig:asy_push} (left) illustrates the procedure for the blue
agent $i$.
Candidate vertices are considered in distance-based order, with
\flip (Sec. \ref{method:flip}) optionally breaking ties.
For the candidate $v_1$, the blue agent moves immediately if the
vertex is not occupied.
If $v_1$ is occupied by an unplanned red agent $j$, the procedure
attempts to push it away by recursively calling \textsc{AsyPush} on $j$. The recursive call may push other unplanned agents.
If the red agent cannot be pushed away, the blue agent tries the
next candidate $v_2$.
Vertices reserved by already assigned actions are rejected during push procedure. If red agent can be successfully pushed, the blue agent cannot enter $v_1$ as soon as the red agent starts moving.
Because a moving agent occupies both endpoints of its edge,
$v_1$ becomes available only when the red agent completes its current move, at the release time $t_j^{\mathrm{rel}}$ returned by the recursive call. The blue agent is therefore assigned a current wait ending at
$\tau_{\mathrm{next}}$. Its subsequent move to $v_1$, scheduled
to start at $t_j^{\mathrm{rel}}$, is stored in $\Phi$, as shown by
the dashed ``Store'' arrow. In both successful cases, \textsc{AsyPush} returns the time at which
the blue agent arrives at $v_1$.

Formally, $P$ is the set of agents whose actions are already
assigned in $\partial a_{next}$, $U$ the set of vertices occupied by the agents in $P$, and $H$ maps each vertex occupied by an unplanned
agent other than $i$ to that agent
(lines~\ref{alg:asy_push:P}--\ref{alg:asy_push:H}).
The candidates of $i$ are ordered by \textsc{Order}
(line~\ref{alg:asy_push:C}) and tried in this order
(line~\ref{alg:asy_push:for_v}).
A candidate $v \in U$ is reserved by a planned agent and is skipped
(line~\ref{alg:asy_push:check_U}).
If $v$ is occupied by an unplanned agent $j = H[v]$
(lines~\ref{alg:asy_push:if_dom}--\ref{alg:asy_push:get_j}),
$i$ is first assigned a wait at $v^i$
(line~\ref{alg:asy_push:hold_vi}).
This adds $i$ to $P$ and $v^i$ to $U$ in the recursive call, so that
neither $j$ nor any agent pushed by $j$ can be pushed back into
$v^i$.
Then $j$ is pushed recursively (line~\ref{alg:asy_push:recursive}).
If the push fails, the wait of $i$ is withdrawn and the next
candidate is tried (line~\ref{alg:asy_push:check_bot}).
Otherwise $j$ releases $v$ at $t^{\text{rel}}_j$; $i$ keeps its wait
and stores the deferred move $\text{act}(v^i, v, t^{\text{rel}}_j, \bot)$
in $\Phi^i$, which \textsc{LSRPStep} assigns at the event
$t^{\text{rel}}_j$ (line~\ref{alg:asy_push:update_phi}).
The release time of $i$ is then $t^{\text{rel}}_i = t^{\text{rel}}_j + D(i, v^i, v)$,
the time at which $i$ would arrive at $v$
(line~\ref{alg:asy_push:t_push}).
If $v$ is free, $i$ moves to $v$ at the current event, and
$t^{\text{rel}}_i$ is its arrival time $t_v(\partial a^i_{next})$
(lines~\ref{alg:asy_push:assign_v}--\ref{alg:asy_push:t_free}).
In both cases $t^{\text{rel}}_i$ is returned
(line~\ref{alg:asy_push:return}).
If the caller is the \textsc{AsyPush} call that pushed $i$,
$t^{\text{rel}}_i$ is the time at which $i$ vacates $v^i$ for its
pusher.
If no candidate vertex succeeds, $i$ waits at $v^i$ and returns $\bot$ to
report failure
(lines~\ref{alg:asy_push:wait}--\ref{alg:asy_push:return_bot}).

\textsc{Order} (Algorithm~\ref{alg:order}) builds the candidate list
of $i$.
It starts from the neighbours of $v^i$ (line~\ref{alg:order:C}).
The vertex $v^i$ itself (i.e. waiting) is added only when $i$ is
not being pushed (line~\ref{alg:order:wait}): an agent processed
directly by \textsc{LSRPStep} may wait, but a pushed agent must
leave $v^i$ for its pusher.
The candidates are sorted in ascending order of
$D(i, v^i, v) + h^i(v)$, the duration of the action plus a lower
bound on the remaining time to the goal
(line~\ref{alg:order:sort}).
Two optional rules may then reorder the list.
If LSRP*-SWAP is used and a swap is required for $i$, the list is
reversed (lines~\ref{alg:order:swap}--\ref{alg:order:reverse};
Remark~\ref{rem:swap}).
Otherwise, once a solution has been found ($n^* \neq \emptyset$),
\textsc{Flip} breaks the ties of the sorted list
(lines~\ref{alg:order:if_flip}--\ref{alg:order:tie};
Sec.~\ref{method:flip}).

\subsubsection{Flip}\label{method:flip}
\begin{figure}[tb]
    \centering
    \includegraphics[width=\linewidth]{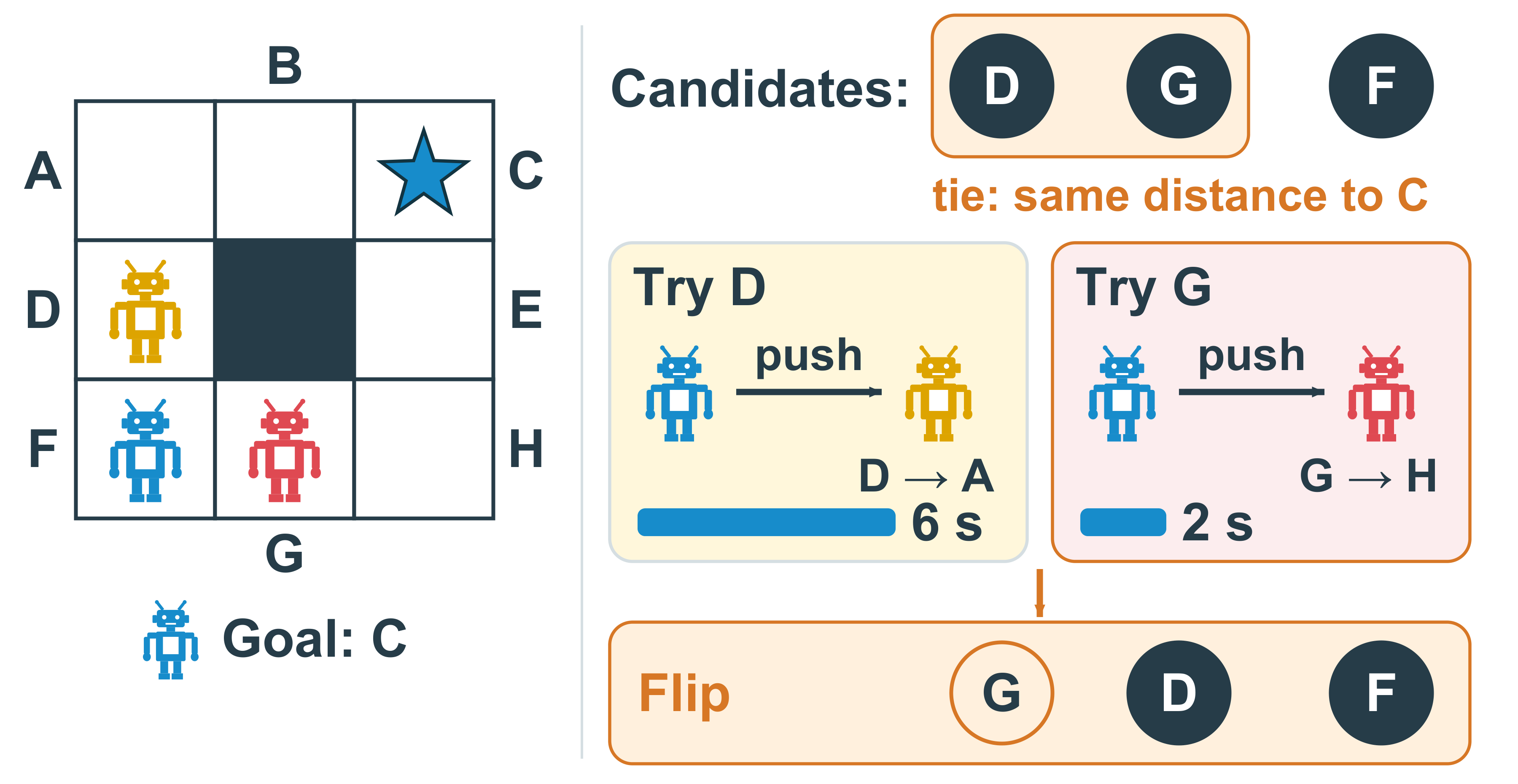}
    \caption{Flip breaks a distance tie using simulated push delays.
    The illustrated delays of $6\,\mathrm{s}$ for $D$ and
    $2\,\mathrm{s}$ for $G$ place $G$ first, while $F$ remains
    a candidate.}
    \label{fig:flip}
\end{figure}
\textsc{Flip}
(lines~\ref{alg:order:if_flip}--\ref{alg:order:tie} of
Algorithm~\ref{alg:order}) is a tie-breaking rule that lets an agent
bypass congestion.
It is applied only after the first solution has been found
($n^* \neq \emptyset$) and only when no swap is triggered.
For each candidate $v$, \textsc{Flip} estimates a delay $\delta(v)$,
which means that how long $i$ would have to wait before it can enter $v$
(lines~\ref{alg:order:for_delay}--\ref{alg:order:delay}).
If $v$ is occupied by an unplanned agent, this agent has to be pushed out of $v$
before $i$ can enter, and $\delta(v)$ estimates when it leaves $v$. This estimate is
computed by \textsc{SimPush}. If the simulated push fails, $\delta(v)$ is set to $0$.
\textsc{SimPush} follows the recursion of \textsc{AsyPush} in a
simplified form: it only reads $\partial a_{next}$, considers moves only ordered by $h$, and returns the estimated time until $H[v]$ releases $v$.
For any other candidate ($v$ free, $v = v^i$, or $v$ reserved by a
planned agent), $\delta(v) = 0$.
The candidates are then stable-sorted by $\delta(v)$ within each tie
of $D(i, v^i, v) + h^i(v)$ (line~\ref{alg:order:tie}), so the
primary order of line~\ref{alg:order:sort} is preserved and
\textsc{Flip} changes only the order among equally good candidates.

Fig.~\ref{fig:flip} exemplifies \textsc{Flip}.
When \textsc{AsyPush} is called for the blue agent, \textsc{Order}
first collects its neighbor vertices $D$, $F$ and $G$ and sorts
them by their distance to the goal $C$.
Vertices $D$ and $G$ have the same distance to the goal.
To break this tie, \textsc{Flip} computes a delay for each of them.
Vertex $D$ is occupied by the yellow agent, and \textsc{SimPush}
estimates that the yellow agent releases $D$ after $6$ seconds.
Similarly, vertex $G$ is occupied by the red agent, which releases
$G$ after $2$ seconds.
Vertex $G$ is therefore placed before $D$, since it is expected to
let the blue agent reach its goal sooner.
\textsc{Flip} only reorders the candidates: $F$ remains in the list
and is tried if neither $G$ nor $D$ can be entered.

\begin{remark}[Swap]\label{rem:swap}
    Pushing alone cannot resolve a head-on conflict in a corridor, where two agents
    keep pushing each other back and forth, so LSRP~\cite{zhou2025loosely} provides a
    swap rule; we write LSRP*-SWAP for LSRP* with this rule.
    A swap is required for $i$ when an adjacent unplanned agent $k$ has to pass $i$
    head-on in a corridor that leads to a vertex where they can pass each other.
    \textsc{Order} then reverses the candidate list
    (lines~\ref{alg:order:swap}--\ref{alg:order:reverse} of Algorithm~\ref{alg:order}),
    so that $i$ first steps back to make room.
    Once $i$ is assigned this move
    (line~\ref{alg:asy_push:assign_v} or~\ref{alg:asy_push:update_phi} of
    Algorithm~\ref{alg:asy_push}), the unplanned $k$ waits at $v^k$ and stores the
    deferred move $\Phi^k \gets \text{act}(v^k, v^i, t^{\text{rel}}_i, \bot)$ into the
    vertex that $i$ releases, so the two agents exchange positions.
    Swap changes only the rule-based completion; the action tree, dominance, and
    re-init are unchanged.
\end{remark}

\section{Analysis}\label{lsrp-star:analysis}\label{lsrp-star-main:analysis}
\begin{lemma}[Retiming]\label{lem:retime}
For search nodes $n,m$ with $e(a(n))=e(a(m))$ and $t_v(a^i(n))\le t_v(a^i(m))$ for all
$i\in I$, every solution $\pi_m$ through $m$ can be retimed into a solution $\pi_n$
through $n$ that performs the same moves after $a(m)$, each ending no later than in
$\pi_m$.
\end{lemma}

\begin{proof}
Let $\pi_m=\langle a_0,\dots,a_K\rangle$ with $a_M=a(m), 0\leq M\leq K$, and let $\hat t$ be the times
in $\pi_n$. We first assume that all actions in $\pi_m$ are moves. For $k>M$, each agent
$i\in I_{curr}(a_{k-1})$ starts a new move $a^i_k$ from $p$ to $v$. Let $a^j_l$,
$M\le l<k$, be the move by which the previous occupant $j$ of $v$ leaves $v$. Since
$\pi_m$ is conflict-free,
\[
t_v(a^i_{k-1})=t_p(a^i_k),\qquad t_v(a^j_l)\le t_p(a^i_k).
\]
If no such $a^j_l$ exists, $v$ is free after $m$, since $e(a(n))=e(a(m))$.
In $\pi_n$, each $a^i_M$ is performed as $a^i(n)$, and for $k>M$, each new
move $a^i_k$ is performed with
\[
\hat t_p(a^i_k)=\max\{\hat t_v(a^i_{k-1}),\,\hat t_v(a^j_l)\},
\]
\[
\hat t_v(a^i_k)=\hat t_p(a^i_k)+\dur{i}{p}{v},
\]
omitting $\hat t_v(a^j_l)$ if $a^j_l$ does not exist; the agent waits in between.
For $k\ge M$, let $P(k)$ denote the statement that $\hat t_v(a^i_l)\le t_v(a^i_l)$ for
all $i\in I$ and $M\le l\le k$, and that these actions are pairwise conflict-free in
$\pi_n$. We prove $P(k)$ by induction on $k$.

\emph{Base case.} $P(M)$ holds, since $t_v(a^i(n))\le t_v(a^i(m))$ for all $i$ by
assumption and $n$ is conflict-free.

\emph{Inductive step.} Assume $P(k-1)$ for some $k>M$, and consider an agent $i$. If
$i\notin I_{curr}(a_{k-1})$, then $a^i_k=a^i_{k-1}$ and the claim follows from
$P(k-1)$. Otherwise, $a^i_k$ is a new move from $p$ to $v$. Since $a^i_{k-1}$ and
$a^j_l$ with $M\le l<k$ are covered by $P(k-1)$,
\[
\hat t_v(a^i_{k-1})\le t_v(a^i_{k-1})=t_p(a^i_k),
\]
\[
\hat t_v(a^j_l)\le t_v(a^j_l)\le t_p(a^i_k),
\]
so $\hat t_p(a^i_k)\le t_p(a^i_k)$ and $\hat t_v(a^i_k)\le t_v(a^i_k)$. Moreover,
$a^i_k$ enters $v$ only after $j$ has left it by $a^j_l$, and every other agent that
occupies $v$ in $a_M,\dots,a_k$ leaves it no later than $j$, so $a^i_k$ causes no
conflict. Hence $P(k)$ holds.

If $\pi_m$ contains waits, the same argument applies with $a^i_{k-1}$ replaced by the
last move of $i$ before $a^i_k$, and the equality
$t_v(a^i_{k-1})=t_p(a^i_k)$ becomes $t_v(a^i_{k-1})\le t_p(a^i_k)$.
\end{proof}

\begin{lemma}[Dominance is sound]\label{lem:dom}
Pruning a dominated node never loses an optimal solution.
\end{lemma}
\begin{proof}
Let $n\dom m$, and let $\pi_m=\langle a_0,\dots,a_K\rangle$ be an optimal solution
through $m$ with $a_M=a(m)$. By Lemma~\ref{lem:retime}, $\pi_m$ can be retimed into a
solution $\pi_n$ through $n$ that performs the same moves after $a(m)$, with
$\hat t_v(a^i_k)\le t_v(a^i_k)$ for all $i\in I$ and $k>M$, where $\hat t$ denotes the
times in $\pi_n$.

Consider an agent $i$ and its action $a^i_K$ in the last joint action, after which
$i$ stays at its goal. If $a^i_K=a^i_M$, i.e., $i$ has no move after $a(m)$, its cost
is $g(a^i(m))$ in $\pi_m$ and $g(a^i(n))$ in $\pi_n$, and $g(a^i(n))\le g(a^i(m))$ by
$n\dom m$. Otherwise, $a^i_K$ is a move after $a(m)$, its cost is $t_v(a^i_K)$ in
$\pi_m$ and $\hat t_v(a^i_K)$ in $\pi_n$, and $\hat t_v(a^i_K)\le t_v(a^i_K)$ by
Lemma~\ref{lem:retime}.

Summing over all agents gives $g(\pi_n)\le g(\pi_m)$, so $\pi_n$ is also optimal.
Hence pruning $m$ loses no optimal solution.
\end{proof}

\begin{lemma}[Wait alignment]\label{lem:wait}
Letting every wait end at the next event $\tau_{next}$ in the action tree does not lose
the optimal solution.
\end{lemma}
\begin{proof}
Consider the search space in which a wait may end at any time. Let $m$ be a search node whose
joint action $a(m)$ contains a wait $a^i(m)$ of agent $i$ at vertex $u$, which ends at
a time $t\neq\tau_{next}$, and let $n$ be the search node that is identical to $m$ except
that this wait ends at $\tau_{next}$.

If $t>\tau_{next}$, both waits traverse the edge $(u,u)$, so $e(a(n))=e(a(m))$.
Moreover, $t_v(a^i(n))<t_v(a^i(m))$ and $g(a^i(n))\le g(a^i(m))$, and the actions of
all other agents are the same. Hence $n\dom m$, and by Lemma~\ref{lem:dom}, pruning $m$
loses no optimal solution.

If $t<\tau_{next}$, no move ends during the wait. Hence no agent releases a vertex during the wait, and the wait can be removed without increasing the cost.

Therefore, letting every wait end at $\tau_{next}$ does not lose the optimal solution.
\end{proof}

\begin{theorem}[Termination]\label{thm:term}
\ours{} generates finitely many search nodes and terminates.
\end{theorem}
\begin{proof}
First, every rule-based call terminates: the only recursion is the push
chain of \textsc{AsyPush}, which operates recursions only on an agent not already
being pushed in the current chain, so its depth is at most $N$.
\begin{itemize}
    \item  \emph{Each search path is finite.} Along a search path, the event time
strictly increases: $\nextevent > \tau(a)$ for joint action $a$. Hence $t_v(a^i)$ is non-decreasing
for every agent $i$, and $g(a^i)$ is non-decreasing because cost is frozen
once $i$ reaches its goal. Thus if two search nodes $n,n'$ ($n$ before $n'$) on the path
share the same joint edge, then $t_v(a^i(n))\le t_v(a^i(n'))$ and
$g(a^i(n))\le g(a^i(n'))$ for all $i$, so $n\dom n'$ and $n'$ is
pruned. If $n$ was itself dominated by another search node $n''$, $n''$ still dominates $n'$ by $n'' \dom n \dom n'$. Therefore each joint edge appears at most once on the search path. Since the workspace is finite, the joint edges are finite, so the search path is finite.

\item \emph{Finitely many search paths.} A search node's action tree has depth at most $N$, and each agent has at most $|\{v^i\}\cup\mathrm{Neigh}(v^i)|$ choices to go, so the action tree has finitely many leaves. Thus every search node has finitely many successors, and thus finite branching.

\end{itemize}
Therefore the search tree has no infinite path, so there are
finitely many search nodes in the search tree. Hence \ours{} terminates.
\end{proof}

\begin{theorem}[Completeness]\label{thm:complete}
    For any solvable MAPF-AA instance, \ours{} returns a solution after generating finitely many search nodes.
\end{theorem}

\begin{proof}
The instance is solvable, so a solution $\pi=(a_0,\dots,a_K)$ exists in
which each $a_k$ contains exactly one move and waits for all other agents,
ending with every agent at its goal. Let $a(n)$ denote the joint action of
a search node $n$.

We prove by induction on $k$ that some search node $n_k$ has
$e(a(n_k))=e(a_k)$. For $k=0$ the root $n_0$ has $e(a(n_0))=e(a_0)$.
Now suppose a search node $n_k$ has $e(a(n_k))=e(a_k)$. Since $a_k$ has one
move and waits otherwise, $n_k$ has the same joint edge, so all agents of
$n_k$ become idle when that move finishes. Because $\pi$ is conflict-free,
$a_{k+1}$ is a valid complete joint action in the action tree $AT(a(n_k))$ of $n_k$. The
search fully expands every action node in $AT(a(n_k))$, so $n_k$ generated a successor $n_{k+1}$ with
$e(a(n_{k+1}))=e(a_{k+1})$. If $n_{k+1}$ is non-dominated, the claim holds; if it
is pruned by dominance, the replacing node has the same joint edge
$e(a_{k+1})$, so the claim holds again.

Hence some search node $n_K$ has $e(a(n_K))=e(a_K)$, whose final vertices
are all goals, i.e.\ $n_K$ is a goal node. Before terminating by Theorem~\ref{thm:term}, the search returns the solution along the path from $n_0$ to $n_K$.
\end{proof}

\begin{theorem}[Optimality]\label{thm:optimal}
If \ours{} terminates with $Open=\emptyset$ on a solvable instance, the
returned solution is optimal.
\end{theorem}

\begin{proof}
Let $h(n)=\sum_i h^i(v(a^i(n)))$ be the admissible heuristic, so $f(n)=g(n)+h(n)$ lower-bounds the cost of every solution through $n$. Let $g^*$ be the optimal cost and $\pi^*=(a_0,\dots,a_K)$ an optimal solution of cost $g^*$. Call a node \emph{unpruned} if it was never removed by dominance or by the $f$-pruned test.

We prove by induction on $t$ the proposition $P(t)$: some unpruned node $n_t$ has $e(a(n_t))=e(a_t)$ and lies on a solution of cost at most $g^*$.

\begin{itemize}
    \item \emph{Base} $t=0$: the root $n_0$ carries $a_0$, so $e(a(n_0))=e(a_0)$; it is
unpruned, and $\pi^*$ (whose first joint action is $a_0$) is a solution of
cost $g^*$ through $a(n_0)$.
\item \emph{Step}: assume $P(t)$. Since $n_t$ lies on a solution of cost at most
$g^*$, $f(n_t)\le g^*$. If the incumbent already has
cost $g^*$, we are done; otherwise $g(n^*)>g^*$. Since the incumbent cost
only decreases, at every earlier time it was at least $g(n^*)>g^*$, so
$f(n_t)\le g^*<g(n^*)$ and $n_t$ was not $f$-pruned when expanded. As $Open$
is empty, the action tree of $n_t$ has been exhausted, so $n_t$ can generate a
successor $n_{t+1}$ with $a(n_{t+1})=a_{t+1}$. Since $a(n_{t+1})$ lies on
$\pi^*$, $f(n_{t+1})\le g^*<g(n^*)$, so $n_{t+1}$ is not $f$-pruned either.
If $n_{t+1}$ is dominated, the replacing node has the same joint edge and,
by Lemma~\ref{lem:dom}, lies on a solution of cost at most $g^*$. Hence $P(t+1)$.
\end{itemize}
At $t=K$, $n_K$ is a goal node of cost at most $g^*$, i.e.\ exactly $g^*$
since $g^*$ is minimal. It is recorded as the incumbent, so the final
incumbent has cost $g^*$. Thus LSRP* reaches the optimal solution when $Open$ is empty.
\end{proof}

\begin{figure}[tb]
  \centering
  \includegraphics[width=\linewidth]{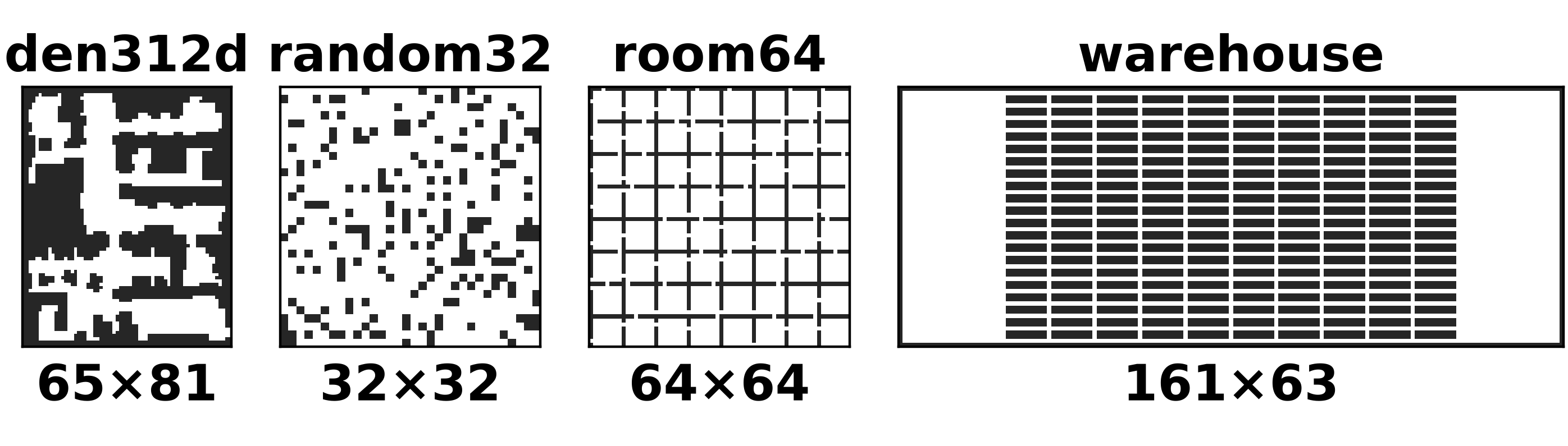}
  \caption{Benchmark maps used in the experiments, with their grid sizes. Obstacles are shown in black.}
  \label{fig:maps}
\end{figure}

\section{Experimental Results}\label{lsrp:result}\label{lsrp-star-main:experiments}
\subsection{Experimental Setup} \label{sec:expe:setup}
We implement LSRP* and LSRP*-SWAP in C++ and run all experiments on a
MacBook Air (Apple M4, 24\,GB RAM).
We use four maps from a MAPF benchmark~\cite{stern2019multi}, shown in
Fig.~\ref{fig:maps}, with 20 instances each and a varying agent number $N$.
They cover different structures:
den312d is a $65\times81$ game map, with irregular open regions connected by narrow passages;
random32 is a $32\times32$ grid with $20\%$ randomly placed obstacles;
room64 is a $64\times64$ grid divided into $8\times8$ rooms of $7\times7$
cells, connected by single-cell doors;
and warehouse is a $161\times63$ grid in which $10\times2$ shelf blocks form
one-cell-wide aisles, with open areas on the left and right sides.
We compare against three baselines: (i) LSRP and LSRP-SWAP~\cite{zhou2025loosely},
using the public implementation with the same random initial priority as ours;
(ii) CBS-AA~\cite{wu2026conflict}, a CBS-based~\cite{sharon2015conflict}
optimal solver for MAPF-AA; and (iii) prioritized planning with
SIPP~\cite{phillips2011sipp}, where we replace SIPP's collision model with
our duration conflict definition (Sec.~\ref{lsrp-star-main:problem-def})
and embed the resulting single-agent SIPP into a prioritized-planning framework for multiple agents. Agent durations are drawn from $\{1,\ldots,5\}$.

\begin{figure}[tb]
  \centering
  \includegraphics[width=.95\columnwidth]{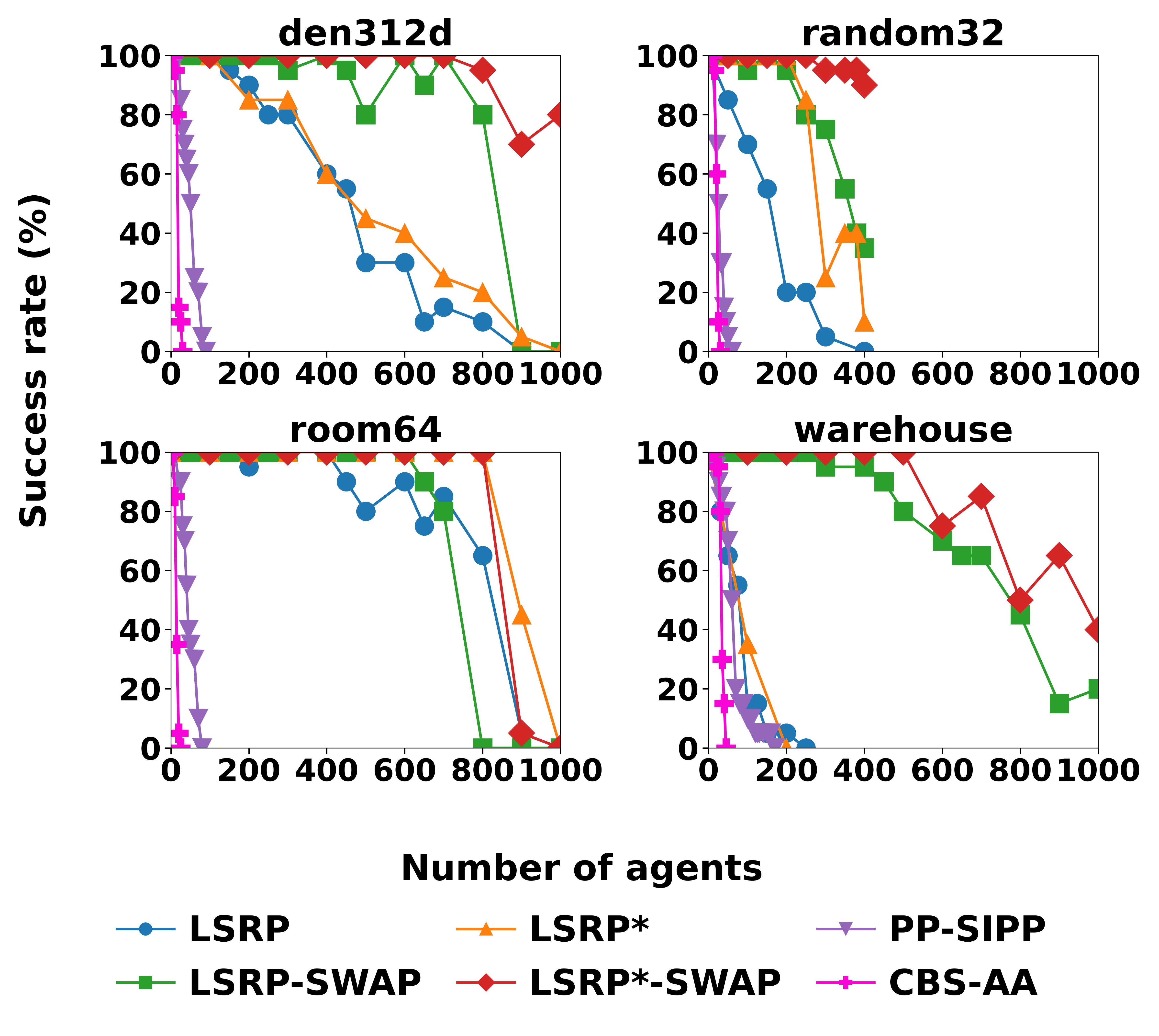}
  \caption{Success rate results. Our methods (LSRP*, LSRP*-SWAP) scale better than the baselines. LSRP*-SWAP scales the best, reaching the upper limit of every benchmark.
}
  \label{fig:success_rate}
\end{figure}

\begin{figure}[tb]
  \centering
  \includegraphics[width=.95\columnwidth]{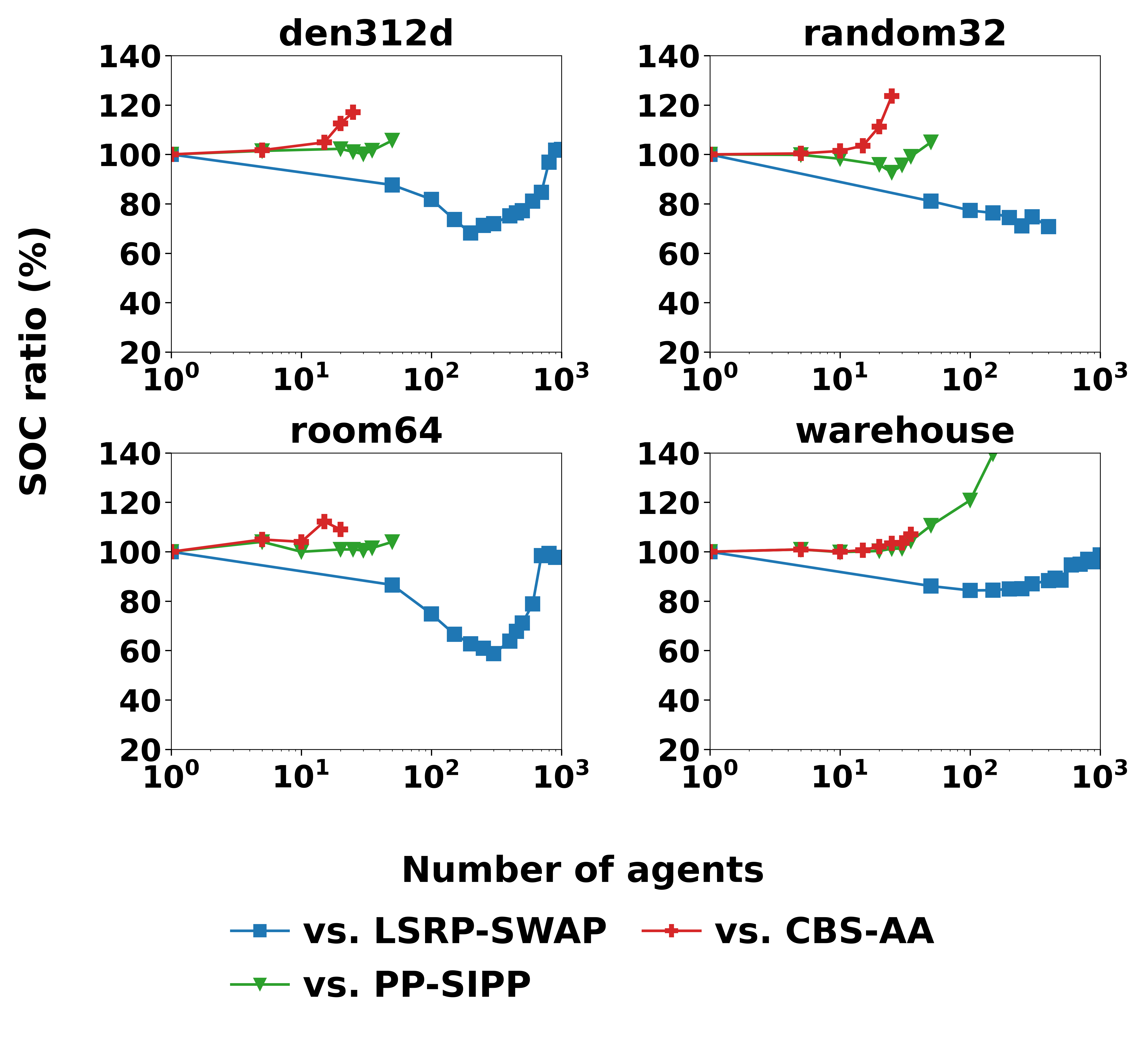}
  \caption{SOC ratio of LSRP*-SWAP over a baseline, computed on instances solved by both.
  The baseline is LSRP-SWAP, CBS-AA, or PP-SIPP. LSRP*-SWAP finds solutions up to about 40\% cheaper than LSRP-SWAP and stays within about 25\% of the optimum.
}
  \label{fig:ratio_60s}
\end{figure}

\subsection{Success Rates} \label{sec:expe:success rate}
With a 30\,s time limit per instance, we vary $N$ from 1 to each benchmark's maximum (Fig.~\ref{fig:success_rate}).
LSRP* and LSRP*-SWAP scale better than the existing approaches; LSRP*-SWAP scales the best, reaching the benchmark's upper limit (400 for Random32, 1000 for the others).
On sparse maps (Room64, Den312d), both LSRP* and LSRP*-SWAP scale well, as there are many cycles in the graph which help resolve conflicts; on cluttered maps (Warehouse, Random32), LSRP* only matches the baselines, whereas LSRP*-SWAP still outperforms all other methods.

\subsection{Solution Quality} \label{sec:expe:solution quality}
To evaluate the refinement capability of LSRP*-SWAP, we examine solution quality using the sum of costs (SOC) of solutions. 
With a 60\,s time limit, we focus on LSRP*-SWAP since it solves more instances than LSRP* at similar cost.
Each plot shows the SOC ratio of LSRP*-SWAP over a baseline; lower is better.
The blue curve (vs.\ LSRP-SWAP) drops well below $100\%$, showing that LSRP*-SWAP finds solutions up to about $40\%$ cheaper than LSRP-SWAP; it then climbs back toward $100\%$ near $N=1000$, where 60\,s no longer suffices for further refinement.
The red curve (vs.\ CBS-AA, optimal but only scalable to small $N$) stays within about $25\%$ of the optimum, showing that LSRP*-SWAP is near-optimal.
The green curve (vs.\ PP-SIPP) is close to $100\%$ on most maps, and rises above it only on the cluttered Warehouse at moderate $N$. It shows LSRP*-SWAP obtains solution quality comparable to PP-SIPP among the instances that were solved by both algorithms.

\begin{figure}[tb]
  \centering
  \includegraphics[width=.95\columnwidth]{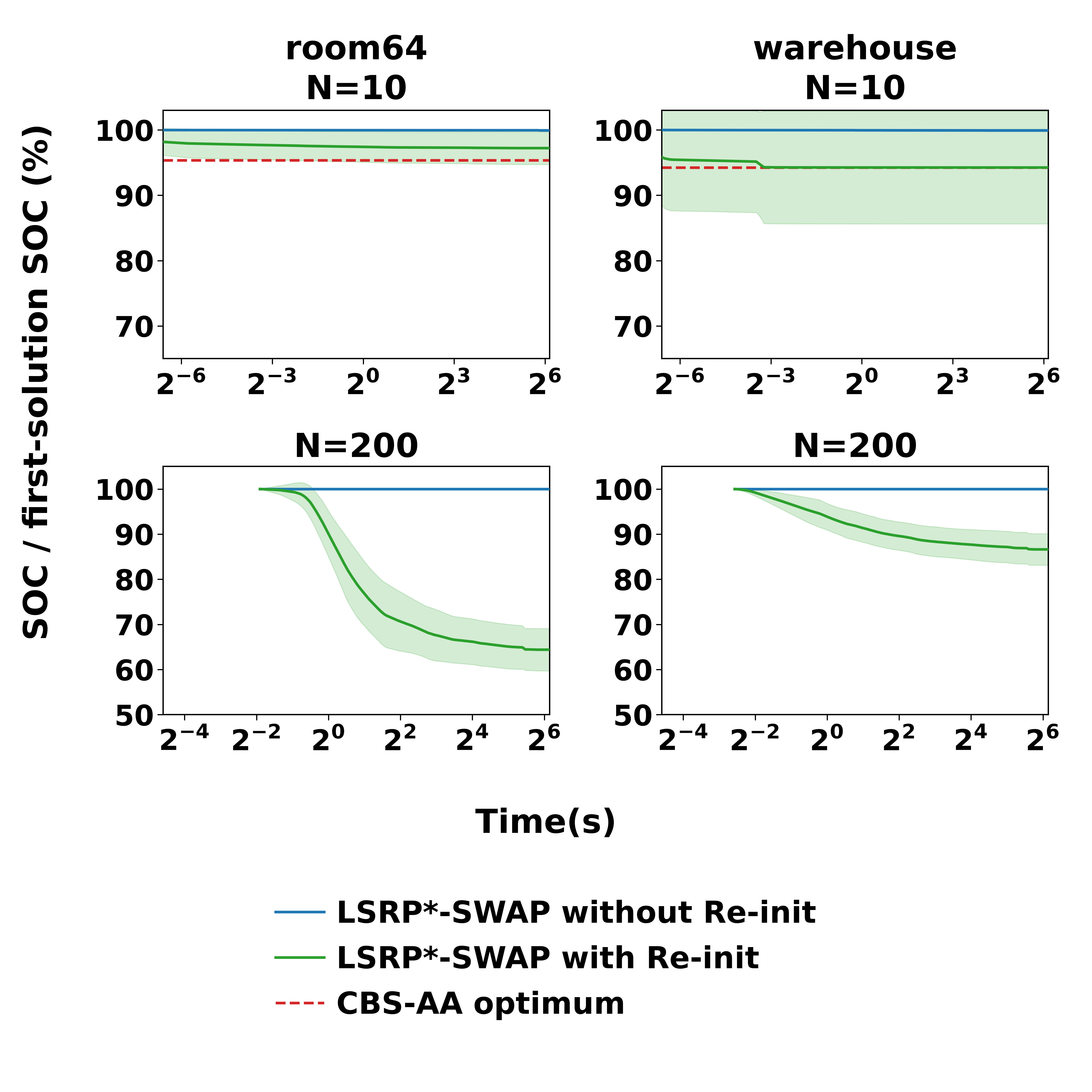}
  \caption{Effect of the proposed re-init on anytime refinement. Solid lines show the
mean SOC normalized by the first-solution SOC over instances, and shaded
bands show $\pm 1$ standard deviation. For $N=10$, the red dashed line
marks the CBS-AA optimum. It verifies that the re-init technique helps improve solution quality over time, especially on large instances ($N = 200$).}
  \label{fig:experiment:reinit}
\end{figure}

\begin{figure}[tb]
  \centering
  \includegraphics[width=.95\columnwidth]{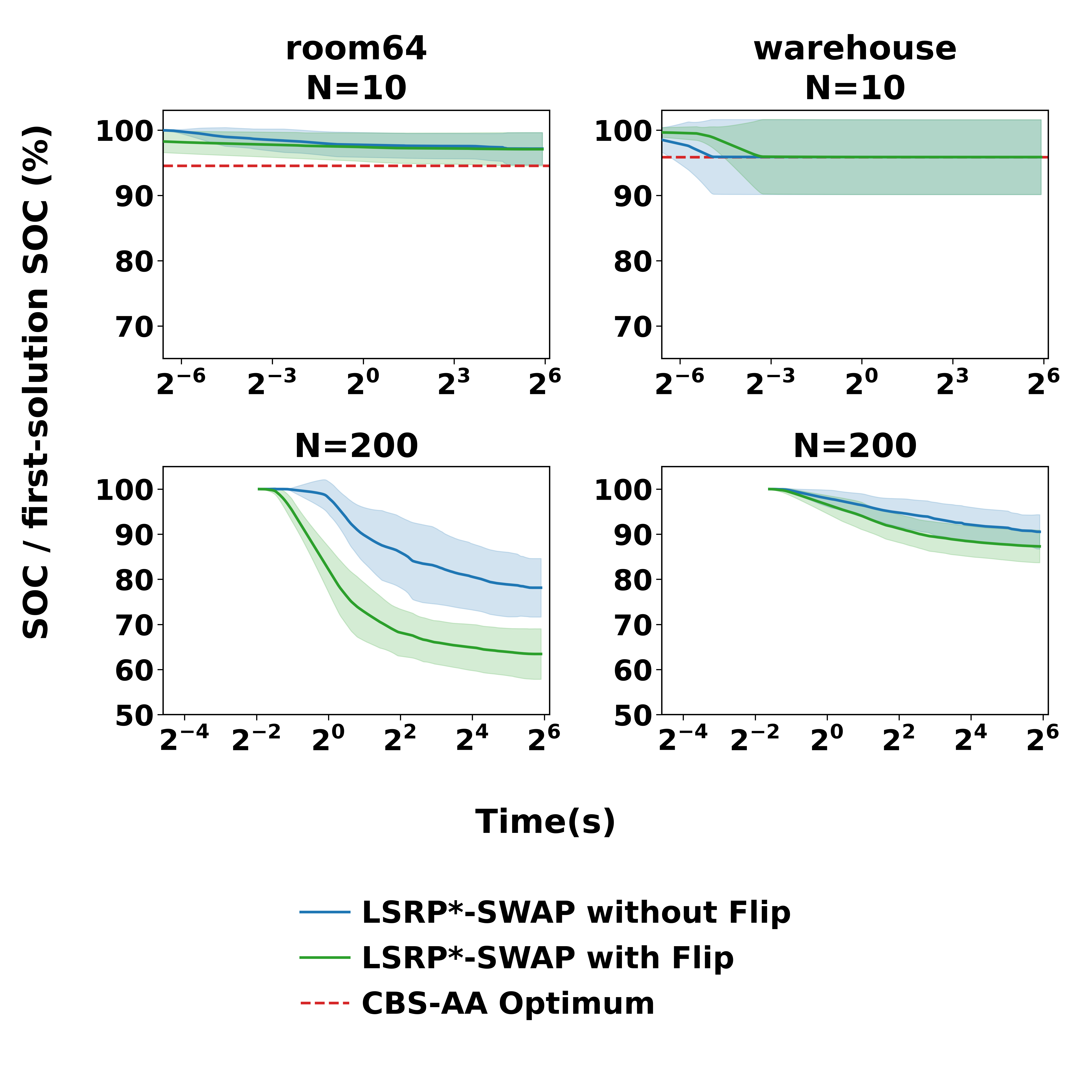}
  \caption{Effect of the proposed Flip technique on anytime refinement. Solid lines show the
mean SOC normalized by the first-solution SOC over instances, and shaded
bands show $\pm 1$ standard deviation. For $N=10$, the red dashed line
marks the CBS-AA optimum. Flip improves solution quality mainly on large instances ($N=200$).}
  \label{fig:flip_results}
\end{figure}

\begin{figure}[tb]
  \centering
  \includegraphics[width=.95\columnwidth]{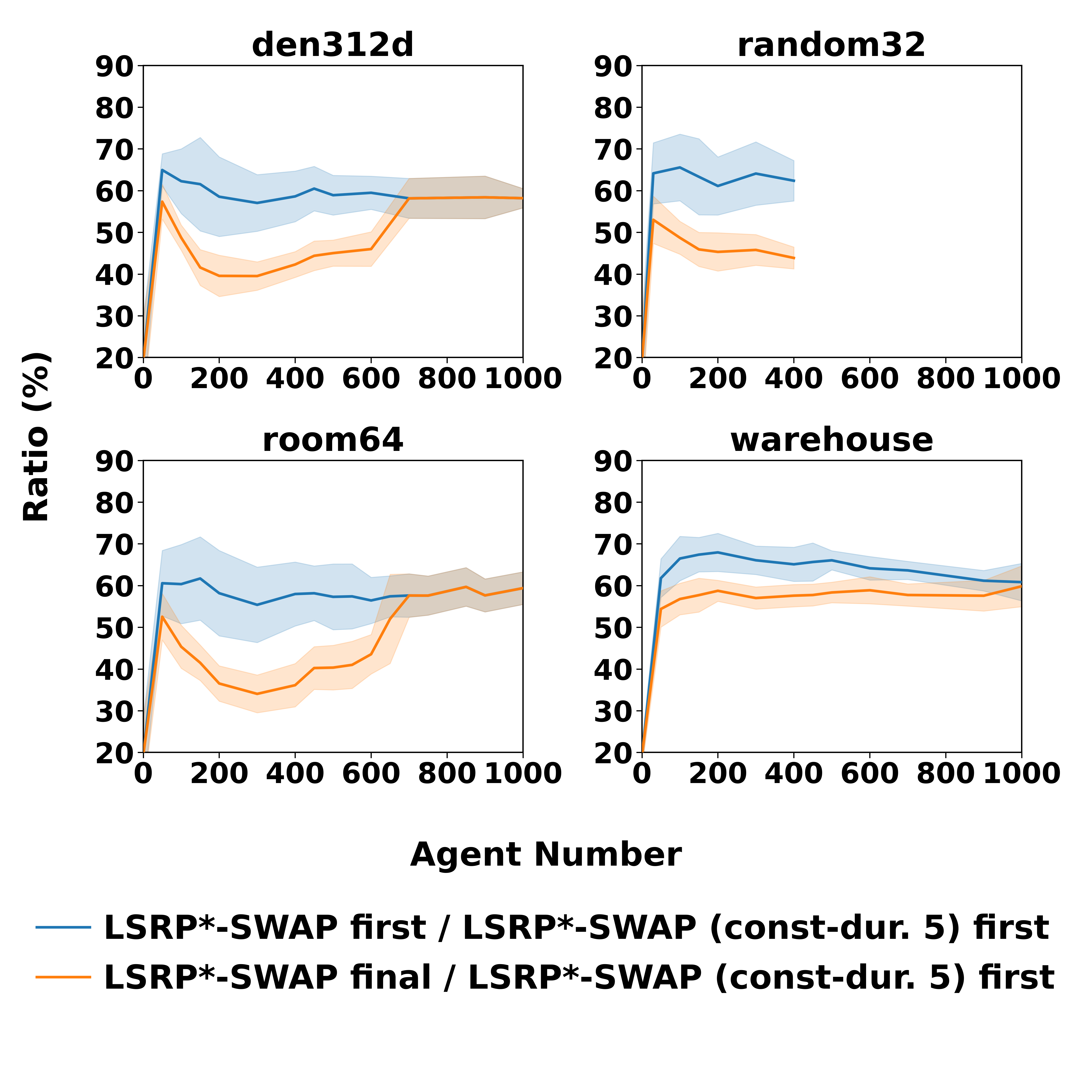}
 \caption{Asynchronous first- and final-solution SOC relative to the
first-solution SOC of the constant-duration-5 baseline. Solid lines
show the mean over instances and shaded bands show $\pm 1$ standard
deviation. Planning with asynchronous actions yields solutions about 35--45\% cheaper than conservative synchronization.
}
  \label{fig:IAA}
\end{figure}

\subsection{Re-init Strategy} \label{sec:expe:reinit}
Under a 60\,s limit, we study re-init on Room64 and Warehouse for $N=10$ and $N=200$; $100\%$ means no improvement over the first solution (Fig.~\ref{fig:experiment:reinit}).
Without re-init, all four curves stay at $100\%$: the search rarely improves the first solution.
With re-init, the green curves drop below $100\%$ over time and continue to decrease, showing steady anytime improvement of the incumbent. The improvement is small at $N=10$ but high at $N=200$.
For $N=10$, it nears optimality ($97\%$ vs.\ CBS-AA $95.4\%$ on Room64; $94\%$ matching CBS-AA on Warehouse); for $N=200$, where LSRP*-SWAP no longer terminates within 60\,s, it cuts SOC by ${\sim}35\%$ on Room64 and ${\sim}13\%$ on Warehouse.
Thus re-init is essential for the anytime refinement of LSRP*-SWAP.

\subsection{Flip Tie-breaking}\label{sec:expe:flip}
We evaluate Flip on Room64 and Warehouse with $N=10$ and $N=200$
under a 60\,s time limit (Fig.~\ref{fig:flip_results}).
We compare LSRP*-SWAP with and without Flip using matched instances.
For each instance, the best-so-far SOC is normalized by its
first-solution SOC before averaging.
The paired runs have identical first-solution costs, so $100\%$
represents their common initial solution quality.
For $N=10$, we additionally restrict the comparison to instances
solved by CBS-AA to obtain a reference on the same instance set.

For $N=10$, the two variants achieve nearly identical final solution
quality.
On Room64, both finish at approximately $97.1\%$ of the first-solution
SOC, above the CBS-AA reference of $94.5\%$.
On Warehouse, both finish at approximately $95.9\%$, closely matching
the CBS-AA reference of $95.8\%$.
Thus, Flip provides little improvement on small instances.

The improvement is high at $N=200$.
On Room64, enabling Flip reduces the SOC ratio from $78.1\%$ to $63.4\%$.
On Warehouse, it reduces the ratio from $90.5\%$ to $87.3\%$.
These results verify that push-delay helps
break ties of distance-ordered candidates and improve
anytime refinement, with a larger benefit on large-scale instances.

\subsection{Impact of Asynchronous Actions} \label{sec:expe:AA}
We further compare asynchronous planning against a conservative synchronized baseline on the instances of Sec.~\ref{sec:expe:solution quality}, both using LSRP*-SWAP with a 60\,s limit.
The asynchronous setting keeps the agent durations in $\{1,\ldots,5\}$, whereas the baseline sets every duration to 5, so that each time slot is long enough for the slowest action and faster agents simply wait.
We normalize the asynchronous first- and final-solution SOC by the mean first-solution SOC of this baseline.
As shown in Fig.~\ref{fig:IAA}, the asynchronous first solution costs about 35--45\% less than the baseline, and anytime refinement makes further improvement to about 35\% on Room64; at larger agent number, the first and final curves converge, as little refinement happens within 60\,s.
These results show the cost of conservative synchronization and the benefit of planning with asynchronous actions.

\section{Conclusion and Future Work}\label{lsrp:conclusion}
This paper develops an anytime planner LSRP* for MAPF-AA by extending LSRP with several newly introduced approaches. The experimental results verify its ability to achieve high scalability for up to a thousand agents across
various maps, while continuing to improve solution quality. Future work includes extending LSRP* from point agents to large agents that occupy
multiple vertices and differ in both size and speed, which requires redefining vertex occupancy and pushing over sets of cells. We also plan to investigate other refinement techniques for MAPF-AA, such as large
neighborhood search, and to validate LSRP* on physical multi-robot platforms with
asynchronous actions.
\bibliographystyle{IEEEtranN} 
\bibliography{references}

@inproceedings{2026_SOCS_BCBS_AA_XuemianWu,
  title = {BCBS-AA: Bounded Sub-Optimal Conflict-Based Search for Multi-Agent Path Finding with Asynchronous Actions},
  author = {Wu, Xuemian and Zhao, Shizhe and Ren, Zhongqiang},
  volume = {},
  doi = {},
  number = {},
  booktitle = {Proceedings of the International Symposium on Combinatorial Search},
  year = {2026},
  month = {},
  pages = {}
}

@inproceedings{phillips2011sipp,
	title={Sipp: Safe interval path planning for dynamic environments},
	author={Phillips, Mike and Likhachev, Maxim},
	booktitle={2011 IEEE International Conference on Robotics and Automation},
	pages={5628--5635},
	year={2011},
	organization={IEEE}
}

@inproceedings{ren2021loosely,
  title={Loosely synchronized search for multi-agent path finding with asynchronous actions},
  author={Ren, Zhongqiang and Rathinam, Sivakumar and Choset, Howie},
  booktitle={2021 IEEE/RSJ International Conference on Intelligent Robots and Systems (IROS)},
  pages={9714--9719},
  year={2021},
  organization={IEEE}
}

@inproceedings{stern2019multi,
  title={Multi-agent pathfinding: Definitions, variants, and benchmarks},
  author={Stern, Roni and Sturtevant, Nathan and Felner, Ariel and Koenig, Sven and Ma, Hang and Walker, Thayne and Li, Jiaoyang and Atzmon, Dor and Cohen, Liron and Kumar, TK and others},
  booktitle={Proceedings of the International Symposium on Combinatorial Search},
  volume={10},
  number={1},
  pages={151--158},
  year={2019}
}

@article{okumura2022priority,
  title = {Priority Inheritance with Backtracking for Iterative Multi-agent Path Finding},
  journal = {Artificial Intelligence},
  pages = {103752},
  year = {2022},
  issn = {0004-3702},
  doi = {https://doi.org/10.1016/j.artint.2022.103752},
  author = {Keisuke Okumura and Manao Machida and Xavier Défago and Yasumasa Tamura},
}

@inproceedings{okumura2023lacam,
  title={Lacam: Search-based algorithm for quick multi-agent pathfinding},
  author={Okumura, Keisuke},
  booktitle={Proceedings of the AAAI Conference on Artificial Intelligence},
  volume={37},
  number={10},
  pages={11655--11662},
  year={2023}
}

@article{okumura2023improving,
  title={Improving lacam for scalable eventually optimal multi-agent pathfinding},
  author={Okumura, Keisuke},
  journal={arXiv preprint arXiv:2305.03632},
  year={2023}
}

@article{ANDREYCHUK2022103662,
title = {Multi-agent pathfinding with continuous time},
journal = {Artificial Intelligence},
volume = {305},
pages = {103662},
year = {2022},
issn = {0004-3702},
doi = {https://doi.org/10.1016/j.artint.2022.103662},
url = {https://www.sciencedirect.com/science/article/pii/S0004370222000029},
author = {Anton Andreychuk and Konstantin Yakovlev and Pavel Surynek and Dor Atzmon and Roni Stern}
}

@article{wagner2015subdimensional,
  title={Subdimensional expansion for multirobot path planning},
  author={Wagner, Glenn and Choset, Howie},
  journal={Artificial intelligence},
  volume={219},
  pages={1--24},
  year={2015},
  publisher={Elsevier}
}

@article{sharon2015conflict,
  title={Conflict-based search for optimal multi-agent pathfinding},
  author={Sharon, Guni and Stern, Roni and Felner, Ariel and Sturtevant, Nathan R},
  journal={Artificial intelligence},
  volume={219},
  pages={40--66},
  year={2015},
  publisher={Elsevier}
}

@inproceedings{de2013push,
  title={Push and rotate: cooperative multi-agent path planning},
  author={De Wilde, Boris and Ter Mors, Adriaan W and Witteveen, Cees},
  booktitle={Proceedings of the 2013 international conference on Autonomous agents and multi-agent systems},
  pages={87--94},
  year={2013}
}

@inproceedings{yu2013structure,
  title={Structure and intractability of optimal multi-robot path planning on graphs},
  author={Yu, Jingjin and LaValle, Steven},
  booktitle={Proceedings of the AAAI Conference on Artificial Intelligence},
  volume={27},
  number={1},
  pages={1443--1449},
  year={2013}
}

@inproceedings{walker2018extended,
  title={Extended Increasing Cost Tree Search for Non-Unit Cost Domains.},
  author={Walker, Thayne T and Sturtevant, Nathan R and Felner, Ariel},
  booktitle={IJCAI},
  pages={534--540},
  year={2018}
}

@inproceedings{barer2014suboptimal,
  title={Suboptimal variants of the conflict-based search algorithm for the multi-agent pathfinding problem},
  author={Barer, Max and Sharon, Guni and Stern, Roni and Felner, Ariel},
  booktitle={Proceedings of the international symposium on combinatorial Search},
  volume={5},
  number={1},
  pages={19--27},
  year={2014}
}

@inproceedings{li2021eecbs,
  title={Eecbs: A bounded-suboptimal search for multi-agent path finding},
  author={Li, Jiaoyang and Ruml, Wheeler and Koenig, Sven},
  booktitle={Proceedings of the AAAI conference on artificial intelligence},
  volume={35},
  number={14},
  pages={12353--12362},
  year={2021}
}

@inproceedings{zhou2025loosely,
  title={Loosely synchronized rule-based planning for multi-agent path finding with asynchronous actions},
  author={Zhou, Shuai and Zhao, Shizhe and Ren, Zhongqiang},
  booktitle={Proceedings of the AAAI conference on artificial intelligence},
  volume={39},
  number={14},
  pages={14763--14770},
  year={2025}
}

@article{wu2026conflict,
  title={Conflict-Based Search for Multi Agent Path Finding with Asynchronous Actions},
  author={Wu, Xuemian and Zhao, Shizhe and Ren, Zhongqiang},
  journal={arXiv preprint arXiv:2603.18866},
  year={2026}
}

@inproceedings{li2021anytime,
  title={Anytime multi-agent path finding via large neighborhood search},
  author={Li, Jiaoyang and Chen, Zhe and Harabor, Daniel and Stuckey, Peter J and Koenig, Sven},
  booktitle={International joint conference on artificial intelligence 2021},
  pages={4127--4135},
  year={2021},
  organization={Association for the Advancement of Artificial Intelligence (AAAI)}
}

@inproceedings{li2022mapf,
  title={MAPF-LNS2: Fast repairing for multi-agent path finding via large neighborhood search},
  author={Li, Jiaoyang and Chen, Zhe and Harabor, Daniel and Stuckey, Peter J and Koenig, Sven},
  booktitle={Proceedings of the AAAI Conference on Artificial Intelligence},
  volume={36},
  number={9},
  pages={10256--10265},
  year={2022}
}

\begin{appendix}\label{app:lss}

\begin{figure}[tb]
  \centering
  \includegraphics[width=\linewidth]{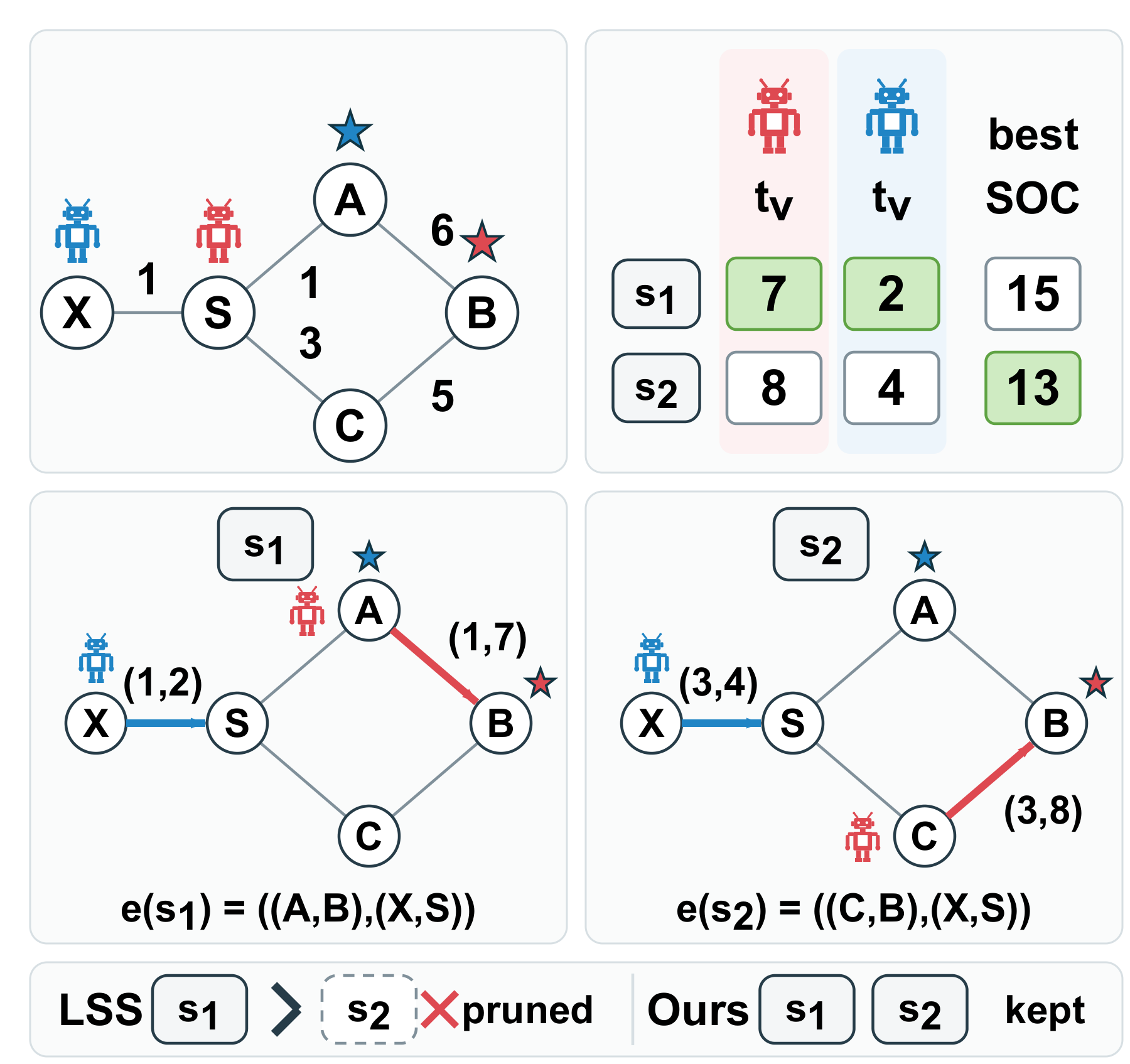}
  \caption{Top left: agent~1 (red) moves from $S$ to $B$ and agent~2 (blue) from $X$ to $A$;
  stars mark goals and edge labels are durations. Bottom: states $s_1$ and $s_2$, with
  $(t_p,t_v)$ next to each move; both have the end joint vertex $(B,S)$ but different joint
  edges. Top right: $s_1$ has smaller arrival times $t_v$, so LSS prunes $s_2$, yet only $s_2$
  leads to the optimal SOC $13$. Definition~1 does not compare them, so both are kept.}
  \label{fig:lss_example}
\end{figure}

Consider the instance in Fig.~\ref{fig:lss_example}. Agent~1 moves from $S$ to $B$
with $\dur{1}{S}{A}=1$, $\dur{1}{A}{B}=6$, $\dur{1}{S}{C}=3$, $\dur{1}{C}{B}=5$, and
agent~2 moves from $X$ to $A$ with $\dur{2}{X}{S}=\dur{2}{S}{A}=1$. Agent~2 must pass
through $S$, so it cannot move before agent~1 leaves $S$.

\emph{Optimal solution.} Agent~1 moves along $(S,C)$ during $[0,3]$ and $(C,B)$ during
$[3,8]$. Agent~2 waits at $X$ until $S$ is released at time $3$, then moves along
$(X,S)$ during $[3,4]$ and $(S,A)$ during $[4,5]$. The SOC is $8+5=13$. This is the
only solution with SOC $13$, and it contains the state $s_2$ at the end joint vertex
$(B,S)$ in which agent~1 moves along $(C,B)$ and agent~2 along $(X,S)$, with arrival
times $(8,4)$.

\emph{Dominating state.} If agent~1 instead moves along $(S,A)$ during $[0,1]$ and
$(A,B)$ during $[1,7]$, agent~2 can move along $(X,S)$ during $[1,2]$. This gives a
state $s_1$ at the same end joint vertex $(B,S)$, with arrival times $(7,2)$. Since
$7<8$ and $2<4$, $s_1$ strictly dominates $s_2$ under the rule of LSS. However, $s_1$ cannot lead to a solution with
SOC $13$: agent~1 occupies $A$ until time $7$, so agent~2 reaches $A$ no earlier than
$8$ and the SOC is at least $15$. As a result, LSS returns a solution with SOC $14$,
e.g.\ agent~1 waits at $C$ during $[3,4]$, while the optimal SOC is $13$.

Under Definition~\ref{def:dominance}, these two joint states have different joint edges,
$((A,B),(X,S))$ and $((C,B),(X,S))$, so they are not compared and $s_2$ is kept.

\end{appendix}

\end{document}